\documentclass[journal, twocolumn,10pt]{IEEEtran}

\usepackage{multicol}
\usepackage{multirow}
\usepackage{array}

\usepackage[final]{graphicx}
\usepackage{graphics}
\usepackage{cite}
\usepackage{amsmath,amssymb,algorithmic,algorithm}

\usepackage[margin=1in]{geometry}
\usepackage{subfig}
\usepackage[usenames]{color}
\definecolor{plum}  {rgb}{.4,0,.4}
\definecolor{forest}  {rgb}{0,.6,0}
\definecolor{midnight}  {rgb}{0,0,.8}
\usepackage[plainpages=false,pdfpagelabels,colorlinks=true,urlcolor=midnight,linkcolor=midnight,citecolor=forest]{hyperref}
\usepackage{setspace}

\usepackage{soul}
\hypersetup{colorlinks=true}
\DeclareMathOperator{\pen}{pen}
\newcommand{\cM}{\mathcal{M}}

\newcommand{\email}[1]{\texttt{\href{mailto:#1}{#1}}}

\def\ie{{\em{i.e.,~}}}

\newcommand{\wh}[1]{\widehat{#1}}
\newcommand{\Expect}{\mathbb{E}}

\newcommand{\yao}[1]{{\color{black}{#1}}}

\hypersetup{colorlinks=true}

\newcommand{\ben}{\begin{eqnarray}}

\newcommand{\een}{\end{eqnarray}}

\newcommand{\transpose}{^{\top}}
\newcommand{\cP}{\mathcal{P}}
\newtheorem{thm}{Theorem}

\newcommand{\tr}{\mbox{tr}}
\newcommand{\Set}{\mathcal{S}}
\newcommand{\ka}{\kappa}

\title{Change-point detection for high-dimensional \\ time series with
  missing data}

\author{Yao Xie,\thanks{Yao Xie (email: \email{yao.c.xie@gmail.com}),
    Jiaji Huang (email: \email{jiaji.huang@duke.edu}), and Rebecca
    Willett (email: \email{willett@duke.edu}) are with the Department
    of Electrical and Computer Engineering at the Duke
    University.}\quad \and Jiaji Huang, \and \quad Rebecca Willett
  \thanks{This work was supported by DARPA award \# FA8650-11-1-7150,
    AFOSR award \# FA9550-10-1-0390, AFOSR award \# FA9550-11-1-0028,
    and NSF CAREER award \# NSF-CCF-06-43947.}  } \date{\today}

\date{\today}

\begin{document}

\maketitle
\begin{abstract}
  This paper describes a novel approach to change-point detection when
  the observed high-dimensional data may have missing elements.  The
  performance of classical methods for change-point detection typically
  scales poorly with the dimensionality of the data, so that a large
  number of observations are collected after the true change-point
  before it can be reliably detected. Furthermore, missing components
  in the observed data handicap conventional approaches. The proposed
  method addresses these challenges by modeling the dynamic
  distribution underlying the data as lying close to a time-varying
  low-dimensional submanifold embedded within the ambient observation
  space. Specifically, streaming data is used to track a submanifold
  approximation, measure deviations from this approximation, and
  calculate a series of statistics of the deviations for detecting
  when the underlying manifold has changed in a sharp or unexpected
  manner. The approach described in this paper leverages several
  recent results in the field of high-dimensional data analysis,
  including subspace tracking with missing data, multiscale analysis
  techniques for point clouds, online optimization, and change-point
  detection performance analysis. Simulations and experiments
  highlight the robustness and efficacy of the proposed approach in
  detecting an abrupt change in an otherwise slowly varying
  low-dimensional manifold.
\end{abstract}

\section{Introduction}

Change-point detection is a form of anomaly detection where the
anomalies of interest are abrupt temporal changes in a stochastic
process \cite{BassevilleNikiforov1993,PoorHadjiliadis2008}.  A ``quickest''
change-point detection algorithm will accept a streaming sequence of random
variables whose distribution may change abruptly at one time, detect
such a change as soon as possible, and also have long period between
false detections. In many modern applications, the stochastic process
is non-stationary away from the change-points and very high
dimensional, resulting in significant statistical and computational
challenges. For instance, we may wish to quickly identify changes in
network traffic patterns \cite{LakhinaCrovellaDiot2004}, social
network interactions \cite{raginsky_OCP}, surveillance video
\cite{LeeKriegman2005}, \yao{graph structures} \cite{ParkPriebe2012}, or solar flare imagery
\cite{solarFlares,perfectStorm2012}.

Traditional quickest change-point detection methods typically deal with a
sequence of low-dimensional, often scalar, random variables. Na\"ively
applying these approaches to high-dimensional data is impractical
because the underlying high-dimensional distribution cannot be
accurately estimated and used for developing test statistics. This
results in detection delays and false alarm rates that scale poorly
with the dimensionality of the problem. Thus the primary challenge
here is to develop a rigorous method for extracting meaningful
low-dimensional statistics from the high-dimensional data stream
without making restrictive modeling assumptions.

Our method addresses these challenges by using multiscale online
manifold learning to extract univariate change-point detection test
statistics from high-dimensional data. We model the dynamic
distribution underlying the data as lying close to a time-varying,
low-dimensional submanifold embedded within the ambient observation
space. This submanifold model, while non-parametric, allows us to
generate meaningful test statistics for robust and reliable
change-point detection, and the multiscale structure allows for fast,
memory-efficient computations. Furthermore, these statistics can be
calculated even when elements are missing from the observation vectors.


While manifold learning has received significant attention in the
machine learning literature
\cite{isomap,lle,belkinThesis,costaDimension,manifoldRegularization,wakinManifold,AllardChenMaggioni2011,Sofia2012},
online learning of a dynamic manifold remains a significant challenge,
both algorithmically and statistically.  Most existing methods are
``batch'', in that they are designed to process a collection of
independent observations all lying near the same static submanifold,
and all data is available for processing simultaneously.

In contrast, our interest lies with ``online'' algorithms, which
accept streaming data and sequentially update an estimate of the
underlying dynamic submanifold structure, and change-point detection
methods which identify significant changes in the submanifold
structure rapidly and reliably. Recent progress for a very special
case of submanifolds appears in the context of subspace tracking. For
example, the Grassmannian Rank-One Update Subspace Estimation (GROUSE)
\cite{BalzanoNowakRecht2010} and Parallel Estimation and Tracking by
REcursive Least Squares (PETRELS) \cite{ChiEldarCalderbank2012}
\cite{ChiJournal2012} effectively track a single subspace using
incomplete data vectors. The subspace model used in these methods,
however, provides a poor fit to data sampled from a manifold with
non-negligible curvature or a union of \yao{subsets}.

\subsection{Related work}
At its core, our method basically tracks a time-varying probability
distribution underlying the observed data, and uses this distribution
to generate statistics for effective change-point detection.  For
sequential density estimation problems such as this, it is natural to
consider an online kernel density estimation (KDE) method \yao{see, e.g. \cite{PriebeMarchette1993}}. A naive
variant of online KDEs would be quite challenging in our setting,
however, because if we model the density using a kernel at each
observed data point, then the amount of memory and computation
required increases linearly with time and is poorly suited to
large-scale streaming data problems. Ad-hoc ``compression'' or
``kernel herding'' methods for online kernel density estimation
address this challenge \cite{Kristan2010,Kristan2011a} but face computational
hurdles. Furthermore, choosing the kernel bandwidth, and particularly
allowing it to vary spatially and temporally, is a significant
challenge. Recent works consider variable bandwidth selection using
expert strategies which increase memory requirements
\cite{Rigollet07,GrayCake11}. 
Some of these issues are addressed by the RODEO method
\cite{LaffertyDensityRodeo08}, but the sparse additive model assumed
in that work limits the applicability of the approach; our proposed
method is applicable to much broader classes of high-dimensional
densities. Finally, in high-dimensional settings asymmetric kernels
which are not necessarily coordinate-aligned appear
essential for approximating densities on low-dimensional manifolds,
but learning time-varying, spatially-varying, and anisotropic kernels
remains an open problem. In a sense, our approach can be considered a
memory-efficient {\em sparse} online kernel density estimation method,
where we only track a small number of kernels, and we allow the number
of kernels, the center of each kernel, and the shape of each kernel to
adapt to new data over time.

Our approach also has close connections with Gaussian Mixture Models
(GMMs) \cite{MclachlanMixture:00,li,libarron,Goldberger2005}. The basic idea here is
to approximate a probability density with a mixture of Gaussian
distributions, each with its own mean and covariance matrix. The
number of mixture components is typically fixed, limiting the memory
demands of the estimate, and online expectation-maximization
algorithms can be used to track a time-varying density
\cite{mclachlan}. \yao{In the fixed sample-size setting, there has been work reducing the number of components in GMMs while preserving the component structure of the original model \cite{Goldberger2005}.} However, this approach faces several challenges in
our setting. 
In particular, choosing the number of mixture components is
challenging even in batch settings, and the issue is aggravated in
online settings where the ideal number of mixture components may vary
over time. \yao{In the online setting, splitting and merging Gaussian components of an already learned precise GMM has been considered in \cite{Declercq2008}. However, learning a precise GMM online is impractical when data are high-dimensional because, without additional modeling assumptions,
tracking the covariance matrices for each of the mixture components is
very ill-posed in high-dimensional settings.}

Our approach is also closely related to Geometric Multi-Resolution Analysis
(GMRA) \cite{AllardChenMaggioni2011}, which was developed for
analyzing intrinsically low-dimensional point clouds in
high-dimensional spaces. The basic idea of GMRA is to first
iteratively partition a dataset to form a multiscale collection of
subsets of the data, then find a low-rank approximation for the data
in each subset, and finally efficiently encode the difference between
the low-rank approximations at different scales. This approach is a
batch method without a straightforward extension to online settings.

\subsection{Motivating applications}

The proposed method is applicable in a wide variety of
settings. Consider a video surveillance problem.  Many modern sensors
collect massive video streams which cannot be analyzed by human due to
the sheer volume of data; for example, the ARGUS system developed by
BAE Systems is reported to collect video-rate gigapixel imagery
\cite{argus,argus2}, and the Solar Dynamics Observatory (SDO) collects
huge quantities of solar motion imagery ``in multiple wavelengths to
[help solar physicists] link changes in the surface to interior
changes'' \cite{sdo}. Solar flares have a close connection with
geomagnetic storms, which can potentially cause large-scale power-grid
failures. In recent years the sun has entered a phase of intense
activity, which makes monitoring of solar flare bursts an even more
important task \cite{perfectStorm2012}.
With these issues in mind, it is clear that somehow {\em prioritizing}
the available data for detailed expert or expert-system analysis is an essential step in the
timely analysis of such data. If we can reliably detect statistically
significant changes in the video, we can focus analysts' attention on
salient aspects of the dynamic scene. For example, we may wish to
detect a solar flare in a sequence of solar images in real time {\em
  without} an explicit model for flares, or detect anomalous
behaviors in surveillance video
\cite{MahadevanLiBhalodia2010}. Saliency detection has been tackled
previously \cite{Hou07saliencydetection,Seo_staticand}, but most
methods do not track gradual changes in the scene composition and do
not detect {\em temporal} change-points.

A second motivating example is credit history monitoring, where we are
interested in monitoring the spending pattern of a user and raising an
alarm if a user's spending pattern is likely to result a default
\cite{KennedyNamee2011}. Here normal spending patterns may evolve over
time, but we would expect a sharp change in the case of a stolen
identity.

An additional potential application arises in computer network anomaly
detection \cite{Ahmed07multivariateonline}. Malicious attacks or
network failure can significantly affect the characteristics of a
network\cite{PolunchenkoTartakovsky2012, LakhinaCrovellaDiot2004}.
Recent work has shown that network traffic data is well-characterized
using submanifold structure \cite{PatwariHeroPacholski2005}, and using
such models may lead to more rapid detection of change-points with
fewer false alarms.

\subsection{Contributions and paper organization}
The primary contributions of this work are two-fold: we present (a) a
fast method for online tracking of a dynamic submanifold underlying
very high-dimensional noisy data with missing elements and (b) a
principled change-point detection method \yao{using} easily computed
residuals of our online submanifold approximation \yao{based on a sequential generalized likelihood ratio procedure} \cite{SiegmundVenkat1996}. These methods are
supported by both theoretical analyses and numerical experiments on
simulated and real data.

The paper is organized as follows. In Section~\ref{sec:prob} we
formally define our setting and problem. Section~\ref{sec:MOUSSE}
describes our multiscale submanifold model and tracking algorithm,
which is used to generate the statistics used in the change-point
detection component described in Section~\ref{sec:change}. Several
theoretical aspects of the performance of our method are described in
Section~\ref{sec:theory}, and the performance is illustrated in
several numerical examples in Section~\ref{sec:experiments}.


\section{Problem Formulation}\label{sec:prob}

Suppose we are given a sequence of data $x_1, x_2, \ldots$, for
$t = 1, 2, \ldots$, $x_t \in \mathbb{R}^{D}$, where $D$ denotes the
{\em ambient dimension}. The data are noisy measurements of points
lying on a submanifold $\cM_t$: \ben x_t = v_t +
w_t, \qquad \mbox{where} \qquad v_t \in \cM_t. \label{data_model} \een
The {\em intrinsic dimension} of the submanifold $\cM_t$ is $d$. We
assume $d\ll D$. The noise $w_t$ is a zero mean white Gaussian random
vector with covariance matrix $\sigma^2 I$. The underlying submanifold
$\cM_t$ may vary slowly with time.  At each time $t$, we only observe
a partial vector $x_t$ at locations $\Omega_t \yao{\subseteq} \{1, \ldots,
D\}$. Let $\mathcal{P}_{\Omega_t}$ represent the $|\Omega_t|\times D$
matrix that selects the axes of $\mathbb{R}^D$ indexed by $\Omega_t$;
we observe $\mathcal{P}_{\Omega_t} x_t$, where $\Omega_t$ is known.

Our goal is to design an online algorithm that generates a sequence of
approximations $\widehat{\cM}_t$ which track $\cM_t$ when it varies
slowly, and allows us to compute residuals\cite{BassevilleNikiforov1993}  from $\mathcal{M}_t$ for detecting change-points as soon as possible after the submanifold
changes abruptly. The premise is that the statistical properties of
the tracking residuals will be different when the submanifold varies
slowly versus when it changes abruptly.  

Define the operator \ben \mathcal{P}_{\cM} x_t = \arg\min_{x\in\cM}
\|x - x_t\|^2 \label{proj_original} \een as the projection of
observation $x_t$ on to $\cM$, \yao{where $\|x\|$ is the Euclidean norm of a vector $x$}. If we had access to all the data
simultaneously without any memory constraints, we might solve the
following batch optimization problem using all data up to time $t$ for
an approximation:
\begin{align}
  \widehat{\cM}_t^\circ \triangleq \arg\min_{\cM} & \Big\{\sum_{i=1}^t
  \alpha^{t-i} \|\mathcal{P}_{\Omega_i}(x_i -
  \mathcal{P}_{{\cM}} x_i)\|^2  + \mu \pen({\cM}) \Big\},
  \label{M_opt}
\end{align}
where 
$\pen({\cM})$ denotes a regularization term which penalizes the
complexity of $\cM$, $\alpha \in (0, 1]$ is a discounting factor on
the tracking residual  at each time $t$, and $\mu$ is a
user-determined constant that specifies the relative weights of the
data fit and regularization terms. \yao{The cost function in (\ref{M_opt}) is chosen with the following goals in mind: (a) to  balance the tradeoff between tracking residuals and the complexity of our estimator, thereby preventing  over-fitting to data; (b) to track the underlying manifold when it is time-varying via discounting old samples in the cost function; (c) to enable an easy decomposition of cost functions that facilitates online estimation, as we demonstrate in Section \ref{sec:MOUSSE}.}

\begin{figure}[h!]
  \begin{center}
    \includegraphics[width = 0.3\textwidth]{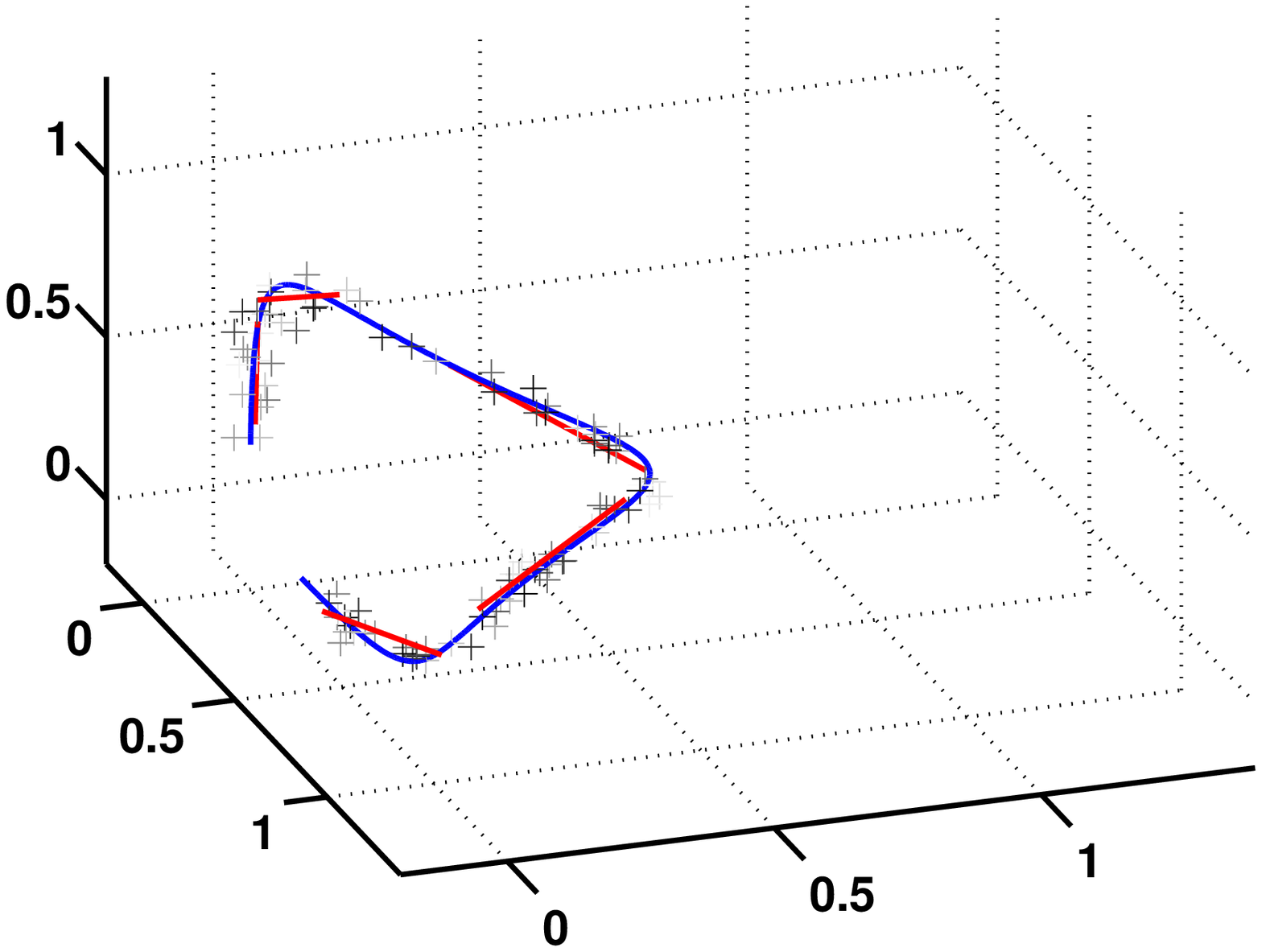}
    \includegraphics[width = 0.3\textwidth]{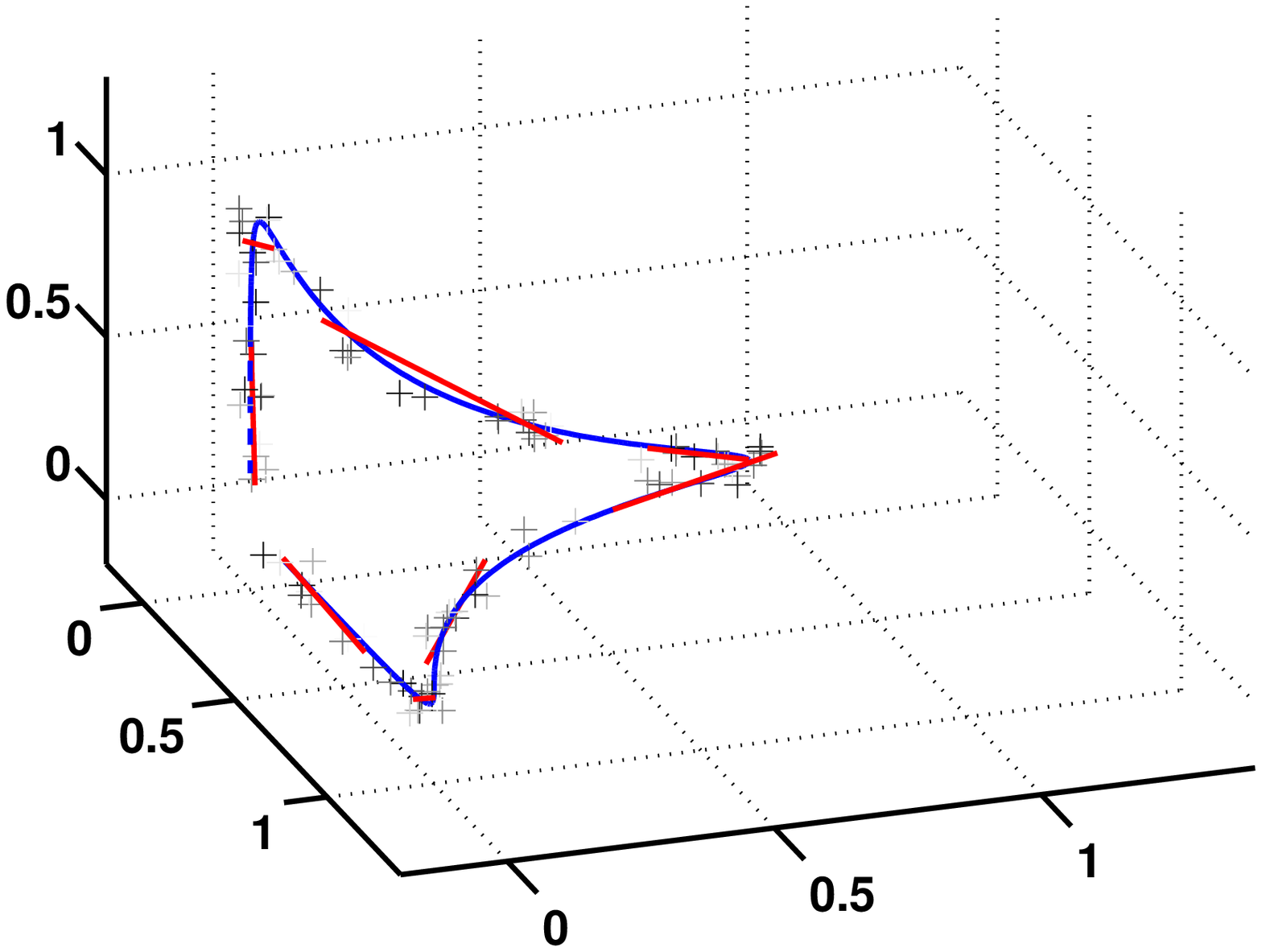}
  \end{center}
  \vspace{0.1in}
  \caption{Approximation of MOUSSE at $t = 250$ (upper) and $t = 1150$
    (lower) of a 100-dimensional submanifold. In this figure we
    project everything into three-dimensional space. The blue curve
    corresponds to true submanifold, the plus signs are noisy samples from
    the submanifold (the lighter plus signs are more dated than the darker
    plus signs), and the red line segments are the approximation subsets computed with
    MOUSSE. As the curvature of the submanifold increases, MOUSSE also
    adapts in the number of subsets.}
  \label{fig:MOUSSE}
\end{figure}

Note that \eqref{M_opt} cannot be solved without retaining all
previous data in memory, which is impractical for the applications of
interest. To address this, 
we instead consider an approximation to the cost function in
\eqref{M_opt} of the form $F(\cM) + \mu \pen(\cM)$, where $F(\cM)$ measures how well the data fits $\cM$. 
In Section~\ref{sec:MOUSSE}, we will show several forms of $F(\cM)$
that lead to recursive updates and efficient tracking algorithms, \yao{and present our new algorithm: Multi-scale Online Union of SubSets Estimate (MOUSSE)}. 
Our method finds a sequence
of approximations $\wh{\cM}_1,\ldots,\wh{\cM}_t$, such that $\wh{\cM}_{t+1}$ is computed by updating the
previous approximation $\wh{\cM}_t$ using $F(\cM)$ and the current
datum $x_{t+1}$ (but not older data). 
 One
example of a MOUSSE approximation is illustrated in
Figure~\ref{fig:MOUSSE}. In this figure, the dashed line corresponds
to the true submanifold, the red lines correspond to the estimated
union of \yao{subsets} by MOUSSE, and the $+$ signs correspond to the past 500
samples, with darker colors corresponding to more recent
observations.  The context is described in more detail in
Section~\ref{sec:track}.

Given the sequence of submanifold estimates
$\wh{\cM}_1,\ldots,\wh{\cM}_t$, we can compute the distance of each
$x_t$ to $\wh{\cM}_t$, which we \yao{refer to as {\em residuals} and denote using} $\{e_t\}$.  We then apply
change-point detection methods to the sequence of tracking residuals
$\{e_t\}$. In particular, we assume that when there is no change-point, the
$e_t$ are i.i.d.\ with distribution $\nu_0$. When there is a change-point,
there exists an unknown time $\ka<t$ such that 
$e_1, \ldots, e_\ka$ are i.i.d.\ with distribution $\nu_0$, and 
$e_{\ka+1}, \ldots$ are i.i.d.\ with distribution
$\nu_1$. Our goals are to (a) detect \yao{as soon as possible} when such a $\kappa$ exists before $t$ and (b) when no such $\kappa$ exists, have our method accept streaming data for as long as possible before falsely declaring a change-point.
(Note that in this setting, even if no change-point exists and all data are i.i.d., any method will eventually incorrectly declare a change-point; that is, for an infinite stream of data, we will have a false alarm at some time with probability one. However, good change-point detection methods ensure that, with high probability, these false detections only occur after a very long waiting time, and thus exert some measure of control over the false alarm rate over time.)

\section{Multiscale Online Union of Subsets Estimation (MOUSSE)}
\label{sec:MOUSSE}

In this section, we describe the Multiscale Online Union of SubSets
Estimation (MOUSSE) method, including the underlying multiscale model
and online update approaches.


\subsection{Multiscale union of \yao{subsets} model}
MOUSSE uses a union of low-dimensional subsets, $\widehat{\cM}_t$, to
approximate $\cM_t$, and organizes these subsets using a tree
structure.
The idea for a multiscale tree structure is drawn from the multiscale
harmonic analysis literature \cite{donoho97cart}. The leaves of the
tree are subsets that are used for the submanifold approximation. Each
node in the tree represents a local approximation to the submanifold
at one scale. The parent nodes are \yao{subsets} that contain coarser
approximations to the submanifold than their children. The subset
associated with a parent node roughly covers the subsets associated
with its two children.

More specifically, our approximation at each time $t$ consists of a
union of \yao{subsets} $\mathcal{S}_{j, k, t}$ that is organized using a
tree structure. Here $j \in\{ 1, \ldots, J_t\}$ denotes the scale or
level of the subset in the tree, where $J_t$ is the tree depth at time
$t$, and $k \in \{ 1,\ldots,2^j\}$ denotes the index of the subset for
that level. The approximation $\widehat{\cM}_t$ at time $t$ is given
by: \ben \widehat{\cM}_t = \bigcup_{(j, k) \in\mathcal{A}_t}
\mathcal{S}_{j, k, t},  \een where $\mathcal{A}_t$ contains the
indices of all {\em leaf} nodes used for approximation at time $t$. Also
define $\mathcal{T}_t$ to be the set of indices of {\em all} nodes in the
tree at time $t$, with $\mathcal{A}_t\subset \mathcal{T}_t.$ 
Each of these subsets lies on a low-dimensional hyperplane with
dimension $d$ and is parameterized as
\begin{equation}
  \begin{split}
    \mathcal{S}_{j, k, t}
    & = \{v \in \mathbb{R}^D:  v = U_{j, k, t} z + c_{j, k, t},\\
    &\qquad \qquad \quad~~ z \transpose \Lambda_{j, k, t}^{-1} z \leq
    1, \quad z \in \mathbb{R}^d\},
  \end{split} \label{subset_expr}
\end{equation}
where the notation $\transpose$ denotes transpose of a matrix or
vector.  The matrix $ U_{j, k, t} \in\mathbb{R}^{D\times d} $ is the
subspace basis, and $c_{j, k, t} \in \mathbb{R}^{D}$ is the offset of
the hyperplane from the origin. The diagonal matrix \[{\Lambda}_{j, k,
  t} \triangleq \mbox{diag}\{\lambda_{j, k, t}^{(1)}, \ldots,
\lambda_{j, k, t}^{(d)}\} \in \mathbb{R}^{d\times d},\] with
$\lambda_{j, k, t}^{(1)}\geq \ldots \geq \lambda_{j, k, t}^{(d)}\geq
0$, contains eigenvalues of the covariance matrix of the projected
data onto each hyperplane.
%
%
%
This parameter specifies the shape of the ellipsoid by capturing the
spread of the data within the hyperplane. In summary, the parameters for
$\mathcal{S}_{j, k, t}$ are \[\{U_{j, k, t}, c_{j, k, t}, \Lambda_{j,
  k, t}\}_{(j, k)\in \mathcal{T}_t},\] and these parameters will be
updated online, \yao{as described in Algorithm \ref{Alg:update}}.

\begin{figure}[h!]
  \centering \centerline{\includegraphics[width=.4\textwidth]{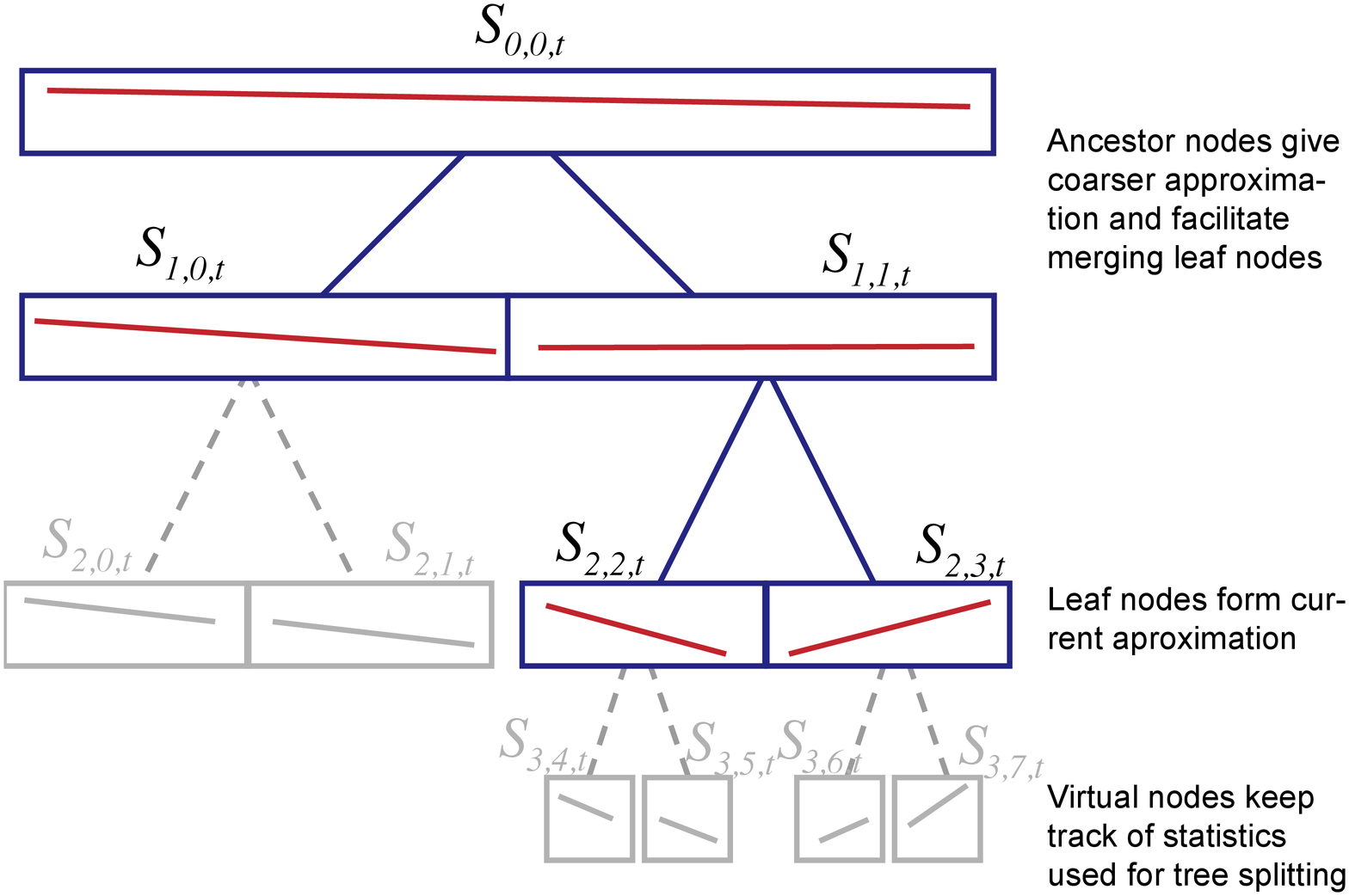}}
  \caption{Illustration of tree structure for \yao{subsets}. The \yao{subsets}
    used in our approximation are $\{\mathcal{S}_{1,0,
      t}\cup\mathcal{S}_{2,2, t}\cup\mathcal{S}_{2,3, t}\}$.}
  \label{Fig:tree}
\end{figure}

In our tree structure, the leaf nodes of the tree also have two {\em
  virtual} children nodes that maintain estimates for corresponding subsets at a finer scale than encapsulated by the leaf nodes of our tree (and $\widehat{\cM}_t$); these subsets are not \yao{used} for our instantaneous submanifold approximation, but rather when 
further subdivision with the tree is needed. \yao{We will explain more details about tree subdivision and growth in Section \ref{sec:tree_update} and Algorithms \ref{alg:split} and  \ref{alg:merge}.} The {\em complexity} of the approximation, denoted $K_t$, is
defined to be the total number of subsets used for approximation at
time $t$:
\begin{equation}
  K_t \triangleq |\mathcal{A}_t|;
\end{equation}
this is used as the complexity regularization term in (\ref{M_opt}):
$ \pen(\widehat{\cM}_t) \triangleq K_t.$ The tree
structure is illustrated in Figure~\ref{Fig:tree}.

\subsection{Approximate Mahalanobis distance}

To update the submanifold approximation, we first determine the
affinity of $x_{t+1}$ to each subset. We might simply project $x_{t+1}$ onto each subset (i.e. ellipsoid), but computing this projection generally requires using numerical solver. Alternatively, we could consider
 the Mahalanobis distance, which is commonly used for data classification and it measures the
quadratic distance of $x$ to a set $\Set$ of data with mean $c =
\Expect\{x\}$ and covariance $\Sigma =
\Expect\{(x-c)(x-c)\transpose\}$. Specifically, the Mahalanobis
distance is defined as
\ben
\varrho(x, \Set) = (x-c)\transpose\Sigma^{-1}(x-c). \label{maha}
\een
However, this distance is only finite and well-defined for points lying in one of the low dimensional subspaces in our approximation. Since our construction is a piecewise linear approximation to a submanifold which may have some curvature, we anticipate many observations which are near but not in our collection of \yao{subsets}, and we need a well-defined, finite distance measure for such points.


To address these challenges, we introduce the {\em approximate Mahalanobis distance} of a point $x$ to a subset $\Set$,  which is a hybrid of Euclidean distance and Mahalanobis distance.  Assume
$x$ with support $\Omega$ and the parameters for
a set $\Set$ is given by $\{U, c, \Lambda\}$. Define
\[
U_{\Omega} \triangleq \mathcal{P}_\Omega U \in
\mathbb{R}^{|\Omega|\times d},  \quad c_{\Omega} \triangleq
\mathcal{P}_\Omega c \in \mathbb{R}^{|\Omega|},
\]
and
\[x_\Omega = \cP_\Omega x \in \mathbb{R}^{|\Omega|}.\]
Define the pseudoinverse operator that computes the coefficients of a
vector in the subspace spanned by $V$ as \ben V^\# \triangleq
(V\transpose V)^{-1}V\transpose.  \een 
Let  $U\transpose_\Omega$ denote  \yao{$(U_\Omega)\transpose,$} and similarly $U^\#_\Omega = (U_\Omega)^\#$.
When $U$ is an orthogonal
matrix, we have $U^\# \equiv U\transpose$, but in general $U_\Omega^\#
\neq U_\Omega\transpose$.  Let
\begin{equation}
  \beta =  U_\Omega^\#   (x_\Omega - c_\Omega), 
  \label{beta}
\end{equation}
and
\begin{equation}
  x_\perp = (I - U_\Omega U_\Omega^\# ) (x_\Omega - c_\Omega).
  \label{beta_perp_def}
\end{equation}
In this definition, $\beta$ is the projection \yao{coefficient} of a re-centered  $x$ on $\yao{U_\Omega}$, and
$x_\perp$ captures the projection residual. 
%
%
Assuming the covariance matrix has a low-rank structure with $d$ large
eigenvalues and $D-d$ small eigenvalues, we can write the
eigendecomposition of the covariance matrix $\Sigma$ as
\[
\Sigma \triangleq \begin{bmatrix}U & U_\perp\end{bmatrix} \Lambda
\begin{bmatrix}U & U_\perp\end{bmatrix}\transpose = U\Lambda_1
U\transpose + U_\perp \Lambda_2 U_\perp\transpose,
\]
where $\Lambda = \mbox{diag}\{\lambda_1, \ldots, \lambda_D\}$,
$\lambda_1\geq\ldots \geq\lambda_D$, $\Lambda_1 =
\mbox{diag}\{\lambda_1, \ldots, \lambda_d\}$, $\Lambda_2 =
\mbox{diag}\{\lambda_{d+1}, \ldots, \lambda_{D}\}$.  If we further
assume that the $D-d$ small eigenvalues are all approximately equal to some 
$\delta > 0$, \ie $\Lambda_2 \approx \delta I$, then the Mahalanobis distance (\ref{maha}) may be approximated as
\ben \varrho(x, \Set)
\approx (x-c)\transpose U\Lambda_1^{-1} U\transpose(x-c) +
\delta^{-1}\|U_\perp \transpose(x-c) \|^2.\label{AM_eg} \een 
\yao{Motivated by this,} we define the  {\em approximate Mahalanobis distance}:  
\begin{equation}
  \rho_\delta(x, \mathcal{S}) 
  \triangleq 
  {\beta}\transpose  {\Lambda}^{-1} {\beta} +  
  \delta^{-1}\|{x}_{\perp}\|^2.
  \label{approx_mahalanobis}
\end{equation}
When the data is complete, $\rho_\delta(x, \Set)$ is equal
to the right-hand-side of (\ref{AM_eg}), since 
\begin{align*}
\beta =& (U\transpose U)^{-1}U\transpose(x-c) = U\transpose (x-c),\\
x_\perp =& (I-UU\transpose)(x-c),
\end{align*}
then we can write the right-hand-side of (\ref{AM_eg}) as $\beta\transpose \Lambda^{-1} \beta + \delta^{-1} \|x_\perp\|^2$. With missing data,
$\rho_\delta(x, \Set)$ is an approximation to $\varrho(x, \Set)$.
%

In definition of the approximation Mahalanobis distance (\ref{approx_mahalanobis}), $\delta$ is a small number and has to be estimated from noisy data. To avoid the numerical instability caused when dividing by a small number, we use the following scaled approximate Mahalanobis distance as a measure of the distance between $x$ and a subset:
\ben
d_\delta(x, \mathcal{S}) = \delta \rho_\delta(x, S) = \delta \beta\transpose \Lambda^{-1} \beta + \|x_\perp\|^2.
\een
With this definition, we can find the subset within our approximation with minimum distance to the new datum $x_t$:
\ben
(j^*, k^*)= \arg\min_{(j, k)} d_{\delta_{j,k,t}}(x_t, \Set_{j, k,
  t}).
  \label{min_S_new}
  \een 
\yao{We can further define the {\em tracking residual} of the submanifold at time $t$.  }
\begin{equation}
\begin{split}
e_t &\triangleq \left( d_{\delta_{j^*,k^*,t}}(x_t, \Set_{j^*, k^*,
  t})\right)^{1/2} \\
&= \left(\delta_{j^*, k^*, t} {\beta^*}\transpose  {\Lambda}_{j^*, k^*, t}^{-1} \beta^* +  
\|{x}_{\perp}^*\|^2\right)^{1/2}, 
\end{split}
\label{e_t_def}
\end{equation}
where $\beta^*$ and $x_\perp^*$ are calculated for $x_{t+1}$ relative to $\mathcal{S}_{j^*, k^*, t}$ using (\ref{beta}) and (\ref{beta_perp_def}). 
\yao{We take the square root of the scaled approximate Mahalanobis distance to ensure that the $e_t$s can be well modeled as draws from a Gaussian distribution (as demonstrated in Section \ref{sec:dist_et}). 

}

\subsection{MOUSSE Algorithm}
When a new sample $x_{t+1}$ becomes available, MOUSSE updates
$\wh{\cM}_t$ to obtain $\wh{\cM}_{t+1}$. The update steps are
presented in Algorithm~\ref{Alg:MOUSSE}; there are three main steps,
detailed in the below subsections: (a) find the subset in 
$\wh{\cM}_t$ which is closest to $x_{t+1}$, (b) update a tracking
estimate of that closest subset, its ancestors, and its nearest virtual child, and (c) grow or
prune the tree structure to preserve a balance between fit to data and
complexity.  \yao{The parameters $\{U_{j, k, t}, \Lambda_{j, k, t}, c_{j,k,t}, \delta_{j, k, t}\}$ are calculated and updated in Algorithm \ref{Alg:update}.} We use $[z]_m$ to denote the $m$-th element of a vector
$z$.

\begin{algorithm}[h!]
  \caption{MOUSSE }
  \begin{algorithmic}[1]
    \STATE Input: \\error tolerance $\epsilon$, step size $\alpha$,
    relative weight $\mu$ \STATE Initialize tree structure, set
    $\epsilon_0 = 0$ \FOR{$t = 0, 1, \ldots$} 
    \STATE Given new data
    $x_{t+1}$ and its support $\Omega_{t+1}$
   %
    \STATE find the minimum
    distance set $\Set_{j^*, k^*, t}$ according to
    \eqref{min_S_new} 
    \STATE let $\beta^*$ and $x_\perp^*$ denote (\ref{beta}) and (\ref{beta_perp_def}) of $x_{t+1}$ for $\Set_{j^*, k^*, t}$
    \STATE calculate: $e_{t+1}$
    using \eqref{e_t_def}
    \STATE update all ancestor nodes and
    closest virtual child node of $(j^*, k^*)$ using
    Algorithm~\ref{Alg:update} \STATE calculate:
    $\epsilon_{t+1} =
    \alpha\epsilon_t + e_{t+1}^2$ \STATE denote parent node
    of $(j^*, k^*)$ as $(j^* - 1, k_p)$ and closest virtual child node as $(j^*+1, k_v)$
    \IF{$\epsilon_{t+1} > \epsilon$ and 
    $d_{\delta_{j^*+1, k_v, t}}(x_{t+1},
      \mathcal{S}_{j^*+1, k_v, t}) + \mu(K_t + 1) < d_{\delta_{j^*, k^*, t}}(x_{t+1},
      \mathcal{S}_{j^*, k^*, t}) +
      \mu K_t$ } \STATE split $(j^*, k^*)$ using
    Algorithm~\ref{alg:split} 
    \ENDIF
    \IF{$\epsilon_{t+1} < \epsilon$ and  $d_{\delta_{j^*-1, k_p, t}}(x_{t+1},
      \mathcal{S}_{j^*-1, k_p, t})+ \mu(K_t - 1) < d_{\delta_{j^*, k^*, t}}(x_{t+1},
      \mathcal{S}_{j^*, k^*, t})  +
      \mu K_t$  } \STATE merge $(j^*, k^*)$ and its sibling using
    Algorithm~\ref{alg:merge} 
    \ENDIF
    \STATE update $\mathcal{A}_t$ and $\mathcal{T}_t$
    \ENDFOR
  \end{algorithmic}
  \label{Alg:MOUSSE}
\end{algorithm}

\begin{algorithm}[h!]
  \caption{Update node}
  \begin{algorithmic}[1]
    \STATE Input: node index $(j, k)$, $\alpha$, $\delta$ and subspace
    parameters \STATE Calculate: ${\beta}$ and $x_\perp$ using
    (\ref{beta}) and (\ref{beta_perp_def}) \STATE Update: $[c_{j, k,
      t+1}]_m = \alpha[c_{j, k, t}]_m + (1-\alpha) [x_{t+1}]_m$,
    $m\in\Omega_{t+1}$ \STATE Update: $\lambda_{j, k, t+1}^{(m)}=
    \alpha\lambda_{j, k, t}^{(m)} + (1-\alpha) [\beta]_{m}^2, m = 1,
    \ldots, d$ \STATE Update: $\delta_{j, k, t+1} = \alpha\delta_{j,
      k, t} + (1-\alpha) \|x_{\perp}\|^2/(D-d)$ \STATE Update
    basis $U_{j, k, t}$ using (modified) subspace tracking algorithm
  \end{algorithmic}
  \label{Alg:update}
\end{algorithm}

%

%
\algsetup{indent=2em}
\begin{algorithm}[h!]
  \caption{Split node $(j, k)$}
  \begin{algorithmic}[1]
    \STATE Turn two virtual children nodes $(j+1, 2k)$ and $(j+1,
    2k+1)$ of node $(j, k)$ into leaf nodes \STATE Initialize virtual
    nodes $(j+1,2k)$ and $(j+1,2k+1)$:
    \begin{align*}
      k_1 &= 2k\\
      k_2 &= 2k+1\\
      c_{j+1, k_1, t+1} &= c_{j, k, t} + \sqrt{\lambda_{j, k,
          t}^{(1)}} u^{(1)}_{j, k, t}/2\\
      c_{j+1, k_2, t+1} &= c_{j, k, t} - \sqrt{\lambda_{j, k,
          t}^{(1)}} u^{(1)}_{j, k,
        t}/2\\
      U_{j+1, k_i, t+1} &= U_{j, k, t}, \quad i = 1, 2 \\
      \lambda_{j+1, k_i, t+1}^{(1)} &= \lambda_{j, k, t}^{(1)}/2, \quad i = 1, 2 \\
      \lambda_{j+1, k_i, t+1}^{(m)} &= \lambda_{j, k, t}^{(m)}, \quad
      m = 2, \ldots, d, \quad i = 1, 2
    \end{align*}
  \end{algorithmic}
  \label{alg:split}
\end{algorithm}

\begin{algorithm}[h!]
  \caption{Merge $(j, k)$ and its sibling}
  \begin{algorithmic}[1]
    \STATE Make the parent node of $(j, k)$ into a leaf node \STATE
    Make $(j, k)$ and its sibling into virtual children nodes of the
    newly created leaf \STATE Delete all four virtual children nodes
    of $(j, k)$ and its sibling
  \end{algorithmic}
  \label{alg:merge}
\end{algorithm}

\subsection{Update subset parameters}\label{sec:tracking}

When updating \yao{subsets}, we can update all \yao{subsets} in our multiscale representation and make the update step-size to be inversely proportional to the approximate Mahalanobis distance between the new sample and each subset, which we refer to as the ``update-all'' approach. Alternatively, we can just update the subset closest to $x_{t+1}$, its virtual children, and all its ancestor nodes, which we refer to as the ``update-nearest'' approach. The update-all approach is computationally more expensive, especially for high dimensional problems, so we focus our attention on the greedy update-nearest approach. The below approaches extend readily to the update-all setting, however.

\yao{In the update-nearest approach,} we update the parameters of the minimum distance subset defined in (\ref{min_S_new}), all its
ancestors in the tree, and its two virtual children. The update
algorithm is summarized in Algorithm~\ref{Alg:update} which denotes
the parameters associated with $\Set_{j^*, k^*, t}$ as $(c, U,
\Lambda, \delta)$, and drops the $j^*$, $k^*$, and $t$ indices for
simplicity of presentation. \yao{The update of the center $c$, $\Lambda$
and $\delta$ are provided in the following, Sections} \ref{sub:c} and \ref{sec:consistency}.

\yao{To decide whether to change the tree structure, we introduce the {\em
  average residual} for a ``forgetting factor'' $\alpha \in (0,1)$:
\begin{equation}
\begin{split}
  \epsilon_t & \triangleq  \sum_{i=1}^t \alpha^{t-i}
e_i^2 \\
  &= \yao{\alpha \epsilon_{t-1} + e_t^2.}
  \end{split}
\end{equation}
We will consider changing the tree
structure when $\epsilon_t$ is greater than our prescribed residual tolerance $\epsilon > 0$.}

Next we will focus on three approaches to updating $U$ \yao{by modifying existing subspace tracking methods. In the following, for tractability reasons, we hold $\Lambda$ fixed and update with respect to $U$ alone at first. We then update the shape parameters $\Lambda$ and $\delta$ for fixed $U$.}

\subsubsection{GROUSE} 
To use GROUSE subspace tracking in this context, we approximate the
first term in (\ref{M_opt}) as
\ben
\begin{split}
  F(\cM)
  =& \sum_{i=1}^t \alpha^{t+1-i} \|\cP_{\Omega_i}(x_i - \cP_{\wh\cM_i} x_i)\|^2  \\
  &+ \|\cP_{\Omega_{t+1}}(x_{t+1}- \cP_{\cM} x_{t+1})\|^2.
\end{split}
\label{GROUSE_method}
\een 
Note the first term is a constant with respect to $\cM$, so we
need only to consider the second term in computing an update. \yao{To focus on updating subspace  without the shape parameters, we replace $ \|\cP_{\Omega_{t+1}}(x_{t+1}- \cP_{\cM} x_{t+1})\|^2$ in (\ref{GROUSE_method}) by 
\ben
f(U)\triangleq \min_a\| \cP_{\Omega_{t+1}} (x_{t+1} - U a - c) \|^2
\label{cost_grouse}
\een (assuming $U$ is orthonormal and including the offset vector
$c$).
The basic idea is now to take a step in the direction of the instantaneous gradient of this cost function (\ref{cost_grouse}). 
%
This task corresponds to the basis update of GROUSE
\cite{BalzanoNowakRecht2010} with the  cost function (\ref{cost_grouse}).} 
%

  Following the same derivation as in
\cite{BalzanoNowakRecht2010}, we have that
\begin{equation}
  \frac{d f}{d U} = -2\mathcal{P}_{\Omega_{t+1}}(x_{t+1} - c - U \beta) \beta\transpose\triangleq -2 r\beta\transpose,
\end{equation}
where $\beta$ is defined in (\ref{beta}), and \[r =
\mathcal{P}_{\Omega_{t+1}}(x_{t+1} - c - U\beta).\] The gradient on
the Grassmannian is given by
\begin{equation}
  \nabla f = (I - UU\transpose)\frac{d f}{dU} = -2(I - UU\transpose)r \beta \transpose = -2 r \beta\transpose,\nonumber
\end{equation}
since $U\transpose r = 0$. We obtain that the update of $U_t$ using
the Grassmannian gradient is given by
\begin{align*}
  U_{t+1} = U_t + \frac{\cos(\xi \eta) -
    1}{\|\beta\|^2}U_t\beta\beta\transpose +
  \sin(\xi\eta)\frac{r}{\|r\|}\frac{\beta\transpose}{\|\beta\|},
\end{align*}
where $\eta > 0$ is the step-size, and $\xi = \|r\|\|U_t\beta\|$. The
step-size $\eta$ is chosen to be $\eta = \eta_0/\|x_{t+1}\|$, for a
constant $\eta_0 > 0$.

\subsubsection{PETRELS} 
Let $(j^*,k^*)$ denote the indices of the closest subset to $x_{t+1}$, and let $\mathcal{I}_t \subseteq \{1,\ldots,t,t+1\}$ denote the set of times corresponding to data which were closest to this subset and used to estimate its parameters in previous rounds. Then we can write
\begin{equation}
\begin{split}
F(\cM) =& \sum_{i \notin \mathcal{I}_t} \alpha^{t-i} \|\cP_{\Omega_i}(x_i - \cP_{\wh\cM_i} x_i)\|^2  \\
&+ \sum_{i \in \mathcal{I}_t} \alpha^{t-i} \|\cP_{\Omega_i}(x_i - \cP_{\cM} x_i)\|^2.
\end{split}
 \label{petrels_opt}
\end{equation}
 where, as before, the first sum is independent of $\cM$ and can be ignored during minimization. \yao{When focusing on updating $U$ for fixed $\Lambda$, }the minimization of $F(\cM)$ with respect to the subspace $U$ used for node $(j^*,k^*)$ in
\eqref{petrels_opt} can be accomplished using the PETRELS
algorithm~\cite{Chi2011}, yielding a solution which can be expressed
recursively as follows. Denoting by $[U]_m$ the $m$-th row of $U$,
we have the update of $U$ given by 
\ben
\begin{split}
  &[U_{t+1}]_m = [U_t]_{m}  \\
  &+ I_{m \in \Omega_t} ([U_t a_{t+1}]_m - a_{t+1}\transpose [U_t]_m
  )(R_{m, t+1})^{\#} a_{t+1},
\end{split}
\een for $ m = 1, \ldots, D$,  where $I_A$ is the indicator function
for event $A$, 
and \[a_{t+1} = (U_t\transpose \cP_{\Omega_{t+1}} U_t)^{\#}
U_t\transpose x_{t+1}.\] The second-order information in $R_{m,t+1}$
can be computed recursively as 
\ben
\begin{split}
 & (R_{m, t+1})^{\#} = \alpha^{-1} (R_{m, t})^{\#} \\ 
   &+\frac{\alpha^{-2}p_{m, t+1}}{1+\alpha^{-1}a_{t+1}\transpose
    (R_{m,t})^{\#} a_{t+1}} (R_{m, t})^{\#} a_t a_t\transpose (R_{m,
    t})^{\#}.
\end{split}
\een

Note that PETRELS does not guarantee the orthogonality of $U_{t+1}$,
which is important for quickly computing projections onto our
submanifold approximation. To obtain orthonormal $U_{t+1}$, we may
apply Gram-Schmidt orthonormalization after each update. We refer to
this modification of PETRELS as {\em PETRELS-GS}. This
orthogonalization requires an extra computational cost on the order of
$\mathcal{O}(Dd^2)$ and may compromise the continuity of $U_t$, \ie
the Frobenius norm $\|U_{t+1}-U_{t}\|_F$ after the orthogonalization
may not be small even when the corresponding \yao{subsets} are very
close~\cite{Meraim2002}. This lack of continuity makes it impossible to effectively track the scale parameter $\Lambda$. %
A faster orthonormalization (FO) strategy with less computation which
also preserves the continuity of $U_t$ is given
in~\cite{Meraim2002}. We refer to this FO strategy combined with
PETRELS as {\em PETRELS-FO}.

\subsubsection{Computational complexity}
For each update with complete data (which is more complex than an update with missing data), the computational complexity of GROUSE is on the order of
$\mathcal{O}(Dd)$, PETRELS-GS is $\mathcal{O}(Dd^2)$, and PETRELS-FO
is $\mathcal{O}(Dd)$.
%
More details about the relative performance of these three subspace
update methods can be found in Section~\ref{sec:experiments}.

\subsection{Tree structure update}\label{sec:tree_update}

When the curvature of the submanifold changes and cannot be
sufficiently characterized by the current subset approximations, we
must perform adaptive model selection. This can be accomplished within
our framework by updating the tree structure -- growing the tree or
pruning the tree, which we refer to as ``splitting'' and ``merging''
branches, respectively. Previous work has derived finite sample bounds
and convergence rates of adaptive model selection in nonparametric
time series prediction \cite{Mei00}.

Splitting tree branches increases the resolution of the approximation
at the cost of higher estimator complexity. Merging reduces resolution
but lowers complexity. When making decisions on splitting or merging,
we take into consideration the approximation residuals as well as the
model complexity (the number of \yao{subsets} $K_t$ used in the
approximation). This is related to complexity-regularized tree
estimation methods \cite{CART,donoho97cart,willett:density} and the
notion of minimum description length (MDL) in compression theory
\cite{BarRisYu98,MerFed98}. In particular, we use the sum of the
average residuals and a penalty proportional to the number of
\yao{subsets} used for approximation as the cost function when deciding to
split or
merge. 
The splitting and merging operations are detailed in Algorithm~\ref{alg:split}
and Algorithm~\ref{alg:merge}. The splitting process mimics the
$k$-means algorithm. In these algorithms, note that for node $(j,k)$
the parent is node $(j-1,\lfloor k/2 \rfloor)$ and the sibling node is
$(j,k+1)$ for $k$ even or $(j,k-1)$ for $k$ odd.

\subsection{Initialization}

To initialize MOUSSE, we assume a small initial training set of
samples, and perform a nested bi-partition of the training data set to
form a tree structure, as shown in Figure~\ref{Fig:tree}. The root of
the tree represents the entire data set, and the children of each node
represent a bipartition of the data in the parent node. The
bipartition of the data can be performed by the $k$-means
algorithm. We start with the entire data, estimate the sample
covariance matrix, perform an eigendecomposition, extract the
$d$-largest eigenvectors and eigenvalues and use them for $U_{1, 1,
  0}$ and $\Lambda_{1, 1, 0}$, respectively. The average of the
$(D-d)$ minor eigenvalues are used for $\delta_{1, 1, 0}$. If the
approximation residual is greater than the prescribed residual tolerance
$\epsilon$, we further partition the data into two clusters using
$k$-means (for $k = 2$) and repeat the above process. We keep
partitioning the data until $\delta_{j, k, 0}$ is less than $\epsilon$
for all leaf nodes. Then we further partition the data one level down
to form the virtual children nodes. This tree construction is similar to that
used in \cite{AllardChenMaggioni2011}.


In principle, it is possible to bypass this training phase and just
initialize the tree with a single root node and two random virtual
children nodes. However, the training phase makes it much easier to
select algorithm parameters such as $\epsilon$ and provides more
meaningful initial virtual nodes, thereby shortening the ``burn in''
time of the algorithm.

\subsection{Choice of parameters}

In general, $\alpha$ should be close to 1, as in the Recursive Least
Squares (RLS) algorithm \cite{adaptiveSigProc}. In the case when the
submanifold changes quickly, we would expect smaller weights for
approximation based on historical data and thus a smaller $\alpha$. In
contrast, a slowly evolving submanifold requires a larger $\alpha$. In
our experiments, $\alpha$ ranges from 0.8 to 0.95. $\epsilon$ controls
\yao{residual tolerance}, which varies from problem to problem according to
the smoothness of the submanifold underlying the data and the noise
variance. Since the tree's complexity is controlled and $\pen({\cM})$
in \eqref{M_opt} is roughly on the order of $\mathcal{O}(1)$, we
usually set $\mu$ close to $\epsilon$.

\section{Change-point detection}\label{sec:change}

We are interested in detecting changes to the submanifold that arise
abruptly and change the statistics of the data. When the submanifold
varies slowly in time, MOUSSE (described in Section~\ref{sec:MOUSSE})
can track the submanifold and produce a sequence of stationary
tracking residuals.  \yao{Because MOUSSE uses a bounded small step-size, and only allows merging or splitting by one level in the tree structure update,} when an abrupt change occurs, MOUSSE \yao{will} lose track of the manifold, resulting in an abrupt increase in the magnitude of the tracking residuals. \yao{This abrupt change in tracking residuals enables change-point detection.} 
\yao{In this section, we formulate the change-point problem using MOUSSE residuals $e_t$, show that the distribution of $e_t$ is close to Gaussian, and adapt the generalized-likelihood ratio (GLR) procedure \cite{SiegmundVenkat1996} for change-point detection.}

\subsection{\yao{Generalized likelihood ratio (GLR)} procedure}

We adopt the quickest change-point detection formulation to detect an abrupt change in the distribution of the residuals. In particular, we
assume that $\nu_0$ is a normal distribution with mean $\mu_0$ and
variance $\sigma_0^2$, and $\nu_1$ is a normal distribution with mean
$\mu_1$ and the same variance $\sigma_0^2$.  Then we can formulate the
change-point detection problem as the following hypothesis test: %
\begin{equation}
  \begin{split}
    \textsf{H}_0: &\quad e_1, \ldots, e_t \sim \mathcal{N}(\mu_0, \sigma_0^2), \\
    \textsf{H}_1: & \quad e_1, \ldots, e_\ka \sim \mathcal{N}(\yao{\mu_0}, \sigma_0^2), \quad e_{\ka+1}, \ldots, e_t \sim \mathcal{N}(\mu_1, \sigma_0^2).
  \end{split}
\end{equation}
\yao{In the case where the pre-change and post-change distributions are completely specified, two very good procedures are the CUSUM test \cite{Page1954, Lorden1971} and the quasi-Bayesian Shiryayev-Roberts procedure \cite{Shiryayev1963, roberts1966} (also see \cite{PoorHadjiliadis2008, changepointArt2012} for surveys). The CUSUM and Shiryayev-Roberts procedures minimize asymptotically to first order the maximum expected delay in detecting a change-point, under different conditions (see \cite{Lorden1971} for CUSUM and \cite{Pollak1985, PollakTartakovsky2009} for Shiryayev-Roberts procedures). }

\yao{In our problem, the post-change distribution is not completely prescribed.} We assume $\mu_0$ and $\sigma_0^2$ are known since typically there is
enough normal data to estimate these parameters (when the training
phase is too short for this to be the case, these quantities can be
estimated online, as described in \cite{PollakSiegmund1991}).
However, we assume $\mu_1$ is unknown since the magnitude of the
change-point can vary from one instance to another. \yao{With this assumption, we instead use the generalized likelihood ratio (GLR) procedure} \cite{SiegmundVenkat1996} (\yao{which is derived based on the CUSUM procedure}), by replacing $\mu_1$ with its maximum likelihood
estimate (for each fixed change-point time $\kappa = k$):
\[
\hat{\mu}_1 = \frac{S_t - S_k}{t-k},
\]
where \[S_t \triangleq \sum_{i=1}^t e_i.\]  \yao{We compute a GLR statistic at
each time $t$ and stop (declare a detected change-point) the first time  the statistic hits a
threshold $b$:}
\begin{align}
  T   = \inf \Big\{&t\geq 1: \max_{t-w\leq k < t}\frac{|(S_t-S_k) -
    \mu_0(t-k)|}{\sigma_0\sqrt{t-k}} \geq b \Big\},\label{det_proc}
\end{align}
where $w$ is a time-window length such that we only consider the most
recent $w$ residuals for change-point detection, and the threshold $b$ is
chosen to control the false-alarm-rate, which is characterized using
average-run-length (ARL) in the change-point detection literature
\cite{Siegmund1985}. Typically we would choose $w$ to be several times
(for example, 5 to 10 times) of the anticipated detection delay, then the
window length will almost have no effect on the detection delay
\cite{XieSiegmund2012}.
This threshold choice is detailed in Section~\ref{sec:arl}.

\subsection{Choice of threshold for change-point detection}
\label{sec:arl}

In accordance with standard change-point detection notation, denote by
$\Expect^\infty$ the expectation when there is no change, \ie
$\Expect_{\textsf{H}_0}$, and by $\Expect^k$ the expectation when
there is a change-point at $\kappa = k$, \ie $\Expect_{\textsf{H}_1,
  \kappa = k}$. The performance metric for a change-point detection
algorithm is typically characterized by the {\em expected detection
  delay} $\sup_{k\geq 0}\mathbb{E}^k\{T-k|T>k\}$ and the {\em
  average-run-length} (ARL) $\mathbb{E}^\infty\{T\}$
\cite{Siegmund1985}. Typically we use $\Expect^0\{T\}$ as a
performance metric since it is an upper bound for $\sup_{k\geq
  0}\mathbb{E}^k\{T-k|T>k\}$.
Note that the \yao{GLR procedure} (\ref{det_proc}) is equivalent to
\begin{equation}
  T = \inf \{t\geq 1: \max_{t-w\leq k < t} 
  \frac{|\tilde{S}_t-\tilde{S}_k|}{\sqrt{t-k}} 
  \geq b,\}
\end{equation}
where $\tilde{S}_t = \sum_{i=1}^t (e_i - \mu_0)/\sigma_0$. Under
$H_0$, we have $(e_i-\mu_0)/\sigma_0$ i.i.d.\ Gaussian distributed
with zero mean and unit variance.  Using the results in
\cite{SiegmundVenkat1996}, we have the following approximation.  When
$b\rightarrow \infty$,
\begin{equation}
  \Expect^\infty \{T\} \sim \frac{(2\pi)^{1/2}\exp\{b^2/2\}}{b\int_0^b x\nu^2(x) dx},
  \label{approx}
\end{equation}
where $\nu(x) = \frac{(2/x)[\Phi(x/2) - 0.5]}{(x/2)\Phi(x/2) +
  \phi(x)/2}$ \cite{XieSiegmund2012}, $\phi(x)$ and $\Phi(x)$ are the
pdf and cdf of the normal random variable with zero mean and unit
variance. We will demonstrate in Section~\ref{sec:ARL_approx} that
this asymptotic approximation is fairly accurate even for finite $b$ \yao{and when $e_t$'s are not exactly Gaussian distributed},
which allows us to choose the change-point detection threshold to
achieve a target ARL without parameter tuning.

\subsection{Distribution of $e_t$}\label{sec:dist_et}

\begin{figure}
  \begin{center}
    \includegraphics[width=0.35\textwidth]{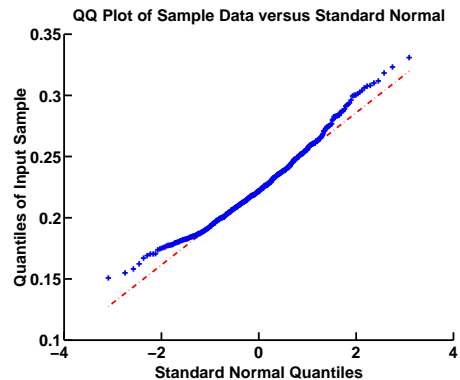}
  \end{center}
  \caption{Q-Q plot of $e_t$, for a $D = 100$ submanifold.}
  \label{QQ}
\end{figure}

In deriving the GLR statistics we have assumed that $e_t$ are
i.i.d.\ Gaussian distributed. A fair question to ask is whether $e_t$
is truly Gaussian distributed, or even to ask whether $e_t$ is a good
statistic to use.
%
%
We can verify that Gaussian distribution is a good approximation for the
distribution of $e_t$ (\ref{e_t_def}).
The QQ-plot of $e_t$ from one of our
numerical examples in Section ~\ref{sec:experiments} when $D = 100$ is
shown in Figure~\ref{QQ}. We will also demonstrate in
Section~\ref{sec:ARL_approx} that the theoretical approximation for
ARL using a Gaussian assumption on $e_t$ is quite accurate.

\section{Performance Analysis}\label{sec:theory}

In this section, we first study the performance of MOUSSE, and then
study the choice for the threshold parameter of the change-point
detection algorithm and provide theoretical approximations. A complete
proof of convergence of MOUSSE (or GROUSE or PETRELS) is challenging since
the space of submanifold approximations we consider is
non-convex. Nevertheless, we can still characterize several aspects of
our approach.

\subsection{MOUSSE residuals}\label{sub:err}

As mentioned earlier, our multiscale subset model is closely related
to geometric multiresolution analysis (GMRA)
\cite{AllardChenMaggioni2011}. In that work, the authors characterize
the favorable approximation capabilities of the proposed multiscale
model. In particular, they prove that the magnitudes of the geometric
wavelet coefficients associated with their algorithm decay
asymptotically as a function of scale, so a collection of data lying
on a smooth submanifold can be well-approximated with a small number
(depending on the submanifold curvature) of relatively large geometric
wavelets. These geometric wavelets are akin to the leaf nodes in our
approximation, so the approximation results of
\cite{AllardChenMaggioni2011} suggest that our model admits accurate
approximations of data on smooth submanifolds with a small number of
leafs.

\subsection{Optimality and consistency}
In Appendix~\ref{sub:c}, we show that the estimate of $c$ is optimal in the complete data setting. In Appendix~\ref{sec:consistency}, we show that the estimates of $\Lambda$ and $\delta$ are consistent in the complete data setting.


\subsection{Missing data}
%
In this section, we show that $\beta$ and $x_\perp$, when using a
missing data projection, are close to their counterparts when using a
complete data projection. Hence, when the fraction of missing data is
not large, the performance of MOUSSE with missing data is also
consistent.
In this section, we omit the subscripts $j$, $k$ and $t$, and denote
$\Omega_t$ by $\Omega$ to simplify notation.
Define the coherence of the basis $U$ as \cite{BalzanoRechtNowak2011}
\ben \textsf{coh}(U) = \frac{D}{d} \max_m \|
UU^\# \textsf{e}_m\|_2^2.  \een
\begin{thm}\label{thm_missing}
  Let $\varepsilon > 0$. Given $x = v + w$, and $w$ is a white
  Gaussian noise with zero mean and covariance matrix $\sigma^2
  I_{D\times D}$. Let $\beta = U\transpose (x - c)$, and $\beta_\Omega
  = U_\Omega^\#(x_\Omega - c_\Omega)$. If for some constant $\ell \in
  (0, 1)$, 
  \begin{equation}
  |\Omega| \geq \max\left\{
    \frac{8}{3}\textsf{coh}(U)d\log(2d/\varepsilon),
    \frac{4}{3}\frac{D}{(1-\ell)\log(2D/\varepsilon)}\right\},
    \label{lower_bound}
    \end{equation}
     then
  with probability at least $1-3\varepsilon$,
  \begin{equation}
    \begin{split}
      \|\beta_\Omega - \beta\|_2^2 \leq& 2\frac{(1+\theta)^2 }{(1-\ell)^2 } \cdot\frac{d}{|\Omega|}\cdot \textsf{coh}(U)\|q\|^2 + \sigma^2\frac{(64/9)D^2}{(1-\ell)^2|\Omega|^2},
    \end{split}
    \label{thm_err_bound}
  \end{equation}
  where \[\theta = \sqrt{2\frac{D\max_{n=1}^D
      |[q]_n|^2}{\|q\|^2}\log(1/\varepsilon)},\] and $q \triangleq (I -
  UU\transpose)(v - c)$.
\end{thm}

\yao{The proof of Theorem \ref{thm_missing} combines techniques from \cite{BalzanoRechtNowak2011} with a new noise bound. Different from \cite{BalzanoRechtNowak2011}, instead of bounding $\|v_\Omega - U_\Omega \beta_\Omega \|$ using $\|v-U\beta\|$, we need to bound $\|\beta - \beta_\Omega\|$ using $\|v-U\beta\|$. The proof of this theorem can be found in Appendix~\ref{app1}. The first term in the lower-bound (\ref{lower_bound}) is a consequence of Lemma 3 in \cite{BalzanoRechtNowak2011}.}
%
%
This theorem shows that the number of non-zero entries, $|\Omega|$,
should be on the order of the maximum of $d\log d$ and  $D/\log(D)$  for
accurate estimation of $\beta_\Omega$. 
%
The first term in the bound (\ref{thm_err_bound}) is proportional to
$\|q\|$,  which is  related to the distance of $v$ from
$U$, and the second term in (\ref{thm_err_bound}) is due to noise.

\section{Numerical Examples}
\label{sec:experiments}

In this section, we present several numerical examples, first based on
simulated data, and then real data, to demonstrate the performance of
MOUSSE in tracking a submanifold and detecting change-points. We also
verify that the theoretical approximation to ARL in Section
\ref{sec:arl} is quite accurate.

\subsection{Comparison of tracking algorithms}

We first compare the performance of different tracking algorithms
presented in Section \ref{sec:tracking}: GROUSE, PETRELS-GS and PETRELS-FO in tracking a time varying manifold. 
The dimension of the submanifold is $D = 100$ and the
intrinsic dimension is $d=1$.  Fixing $\theta \in [-2,2]$, we define
$v(\theta) \in \mathbb{R}^D$ with its $n$-th element
\begin{equation} [v(\theta)]_n=1/\sqrt{2\pi} e^{-(z_n -
    \theta)^2/(2\gamma_t^2)},
  \label{genM}
\end{equation}
where $z_n = -2 + 4n/D$, $n = 1, \ldots, 100$,
corresponds to regularly spaced points between $-2$ and $2$. Let $\gamma_t$ be time-varying: \ben \gamma_t =
\left\{\begin{array}{ll}
    0.6-\gamma_0 t, & t = 1, 2, \dots, s, \\
    0.6-\gamma_0(2s-t), &t = s +1, s+2, \dots,2s,
  \end{array}
\right.  
\label{slow_manifold}
\een where parameter $\gamma_0$ controls how fast the
submanifold changes, and $s=1000$. The observation $x_t$ is obtained from
(\ref{data_model}) with noise variance $\sigma^2 =
4\times10^{-4}$. We compare the methods with various settings of changing rate $\gamma_0$ and percentage of missing entries in $x_t$. 

\begin{figure}[h!]
\begin{center}
\subfloat[$\mathbb{E}\{e_t^2\}$ of MOUSSE using GROUSE]{\label{Fig:grouse}
\includegraphics[width = .4\linewidth]{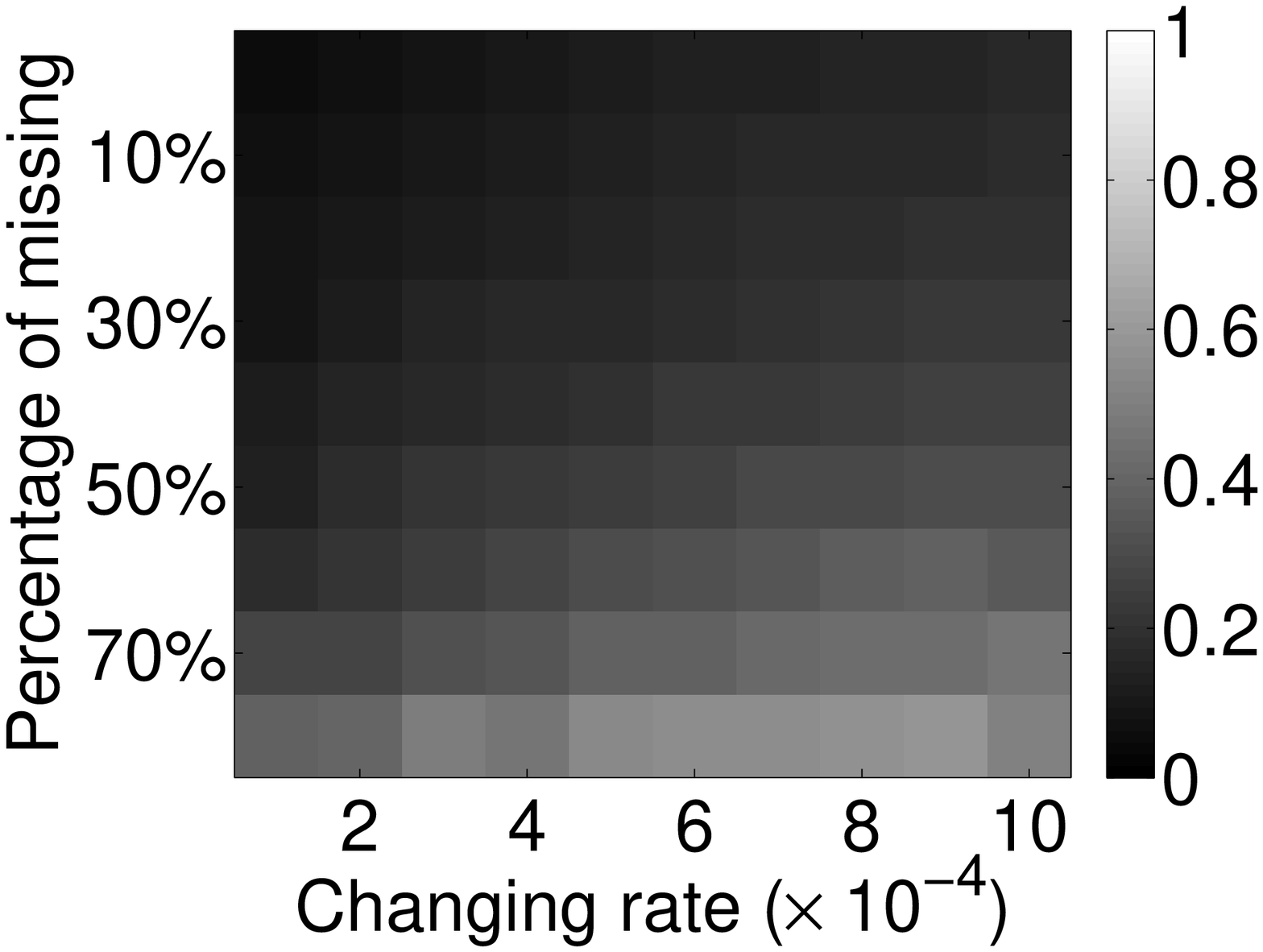}}
\qquad
\subfloat[$\mathbb{E}\{e_t^2\}$ of MOUSSE using PETRELS-GS]
{\label{Fig:petrels}
\includegraphics[width = .4\linewidth]{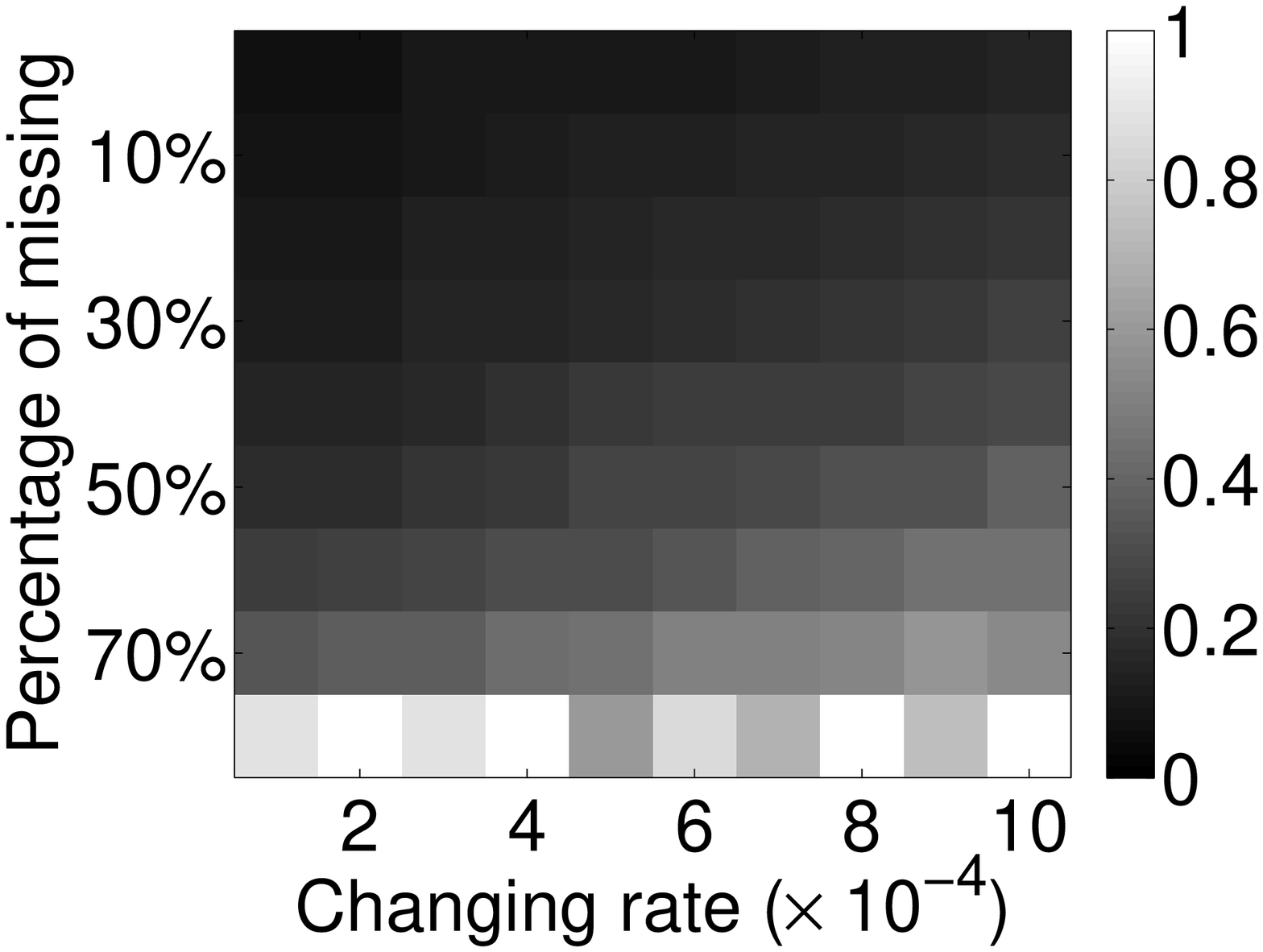}}
\qquad
\subfloat[$\mathbb{E}\{e_t^2\}$ of MOUSSE using PETRELS-FO]{\label{Fig:opast}
\includegraphics[width = .4\linewidth]{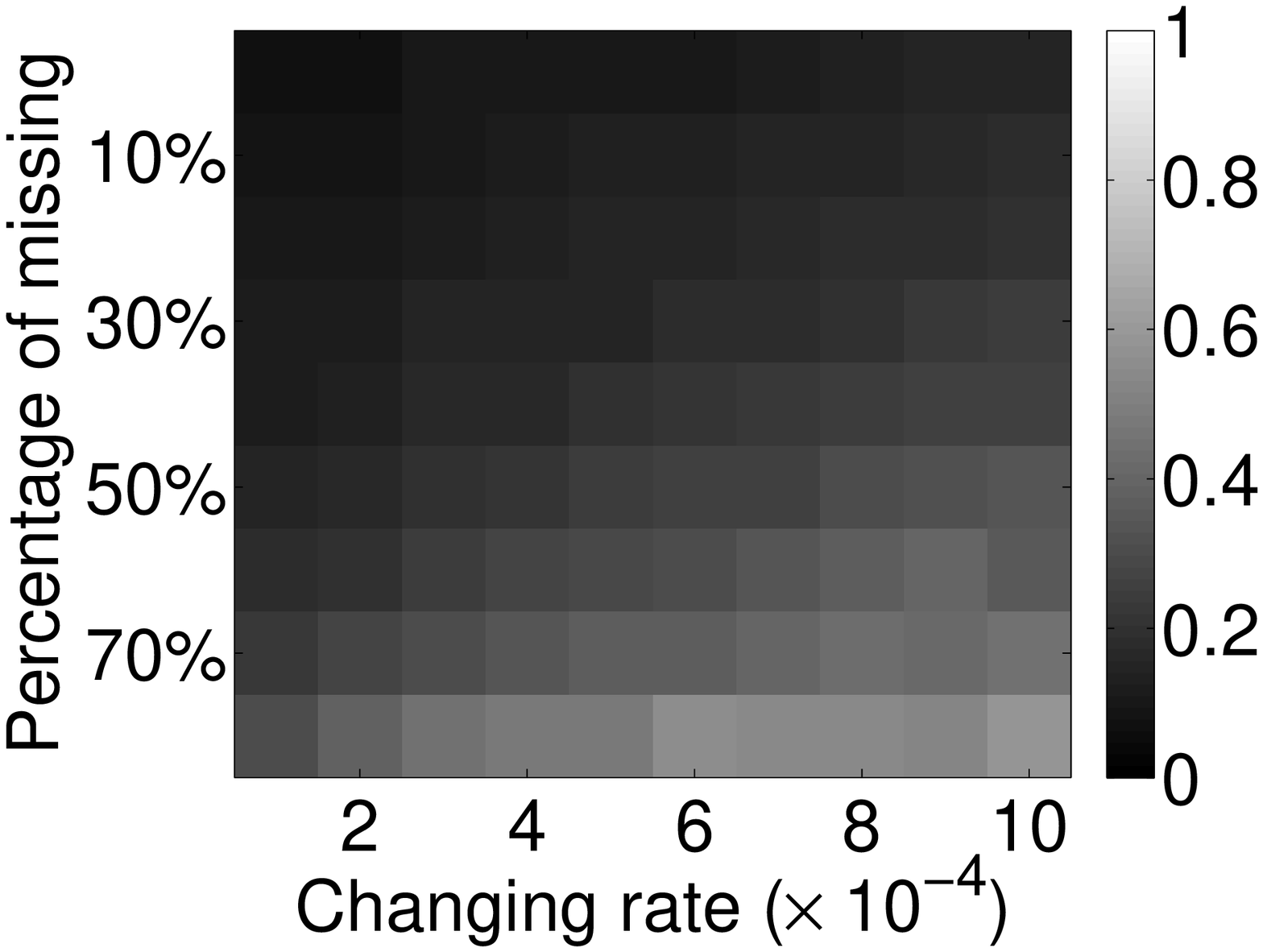}}
\end{center}
\caption{MOUSSE tracking a slowly varying submanifold using: (a) GROUSE,
  (b) PETRELS-GS and (c) PETRELS-FO.  Horizontal axis corresponds to
  rate of change for submanifold and vertical axis corresponds to fraction
  of data missing. Brightness corresponds to $\mathbb{E}\{e_t^2\}$.}
\label{Fig:comp}
\end{figure}

\sloppypar  In the following experiments, we use \yao{sample average approximation error $\varepsilon_N$} obtained from $N = 1200$ samples $\{y_1,\ldots,y_N\}$ as a metric for comparison:
\begin{equation}
\mathbb{E}\{e_t^2\} \approx \frac{1}{N}\sum_{i=1}^N e_i^2,\label{Mc:et}
\end{equation}
where $\Set_{i}$ denotes the minimum distance subset for sample $y_i$. We set the parameters for each tracking algorithm such that they each having the best numerical performance. We use $d = 1$ for MOUSSE in all instances. The comparison results are displayed in
Figure~\ref{Fig:comp}, where the horizontal axis is the submanifold
changing rate $\gamma_0$, the vertical axis is the percentage of missing
data, and the brightness of each block corresponds to our numerical estimate of
$\mathbb{E}\{e_t^2\}$. In Figure~\ref{Fig:comp}, PETRELS-FO performs far better
then PETRELS-GS and slightly better than GROUSE, especially with a large fraction of
missing data. For PETRELS-FO, the best parameters are fairly stable for various combinations of submanifold changing rates and factions of missing data: with
$\alpha$ around 0.9, $\mu$ around 0.2, and $\epsilon$ around 0.1. Considering its lower computational cost and ease of parameter tuning, {\em we adopt PETRELS-FO in MOUSSE for the remaining experiments in this paper.}

\subsection{Tracking a static submanifold}\label{sec:static}

\begin{figure}[h!]
  \begin{center}
    \includegraphics[width = 0.4\textwidth]{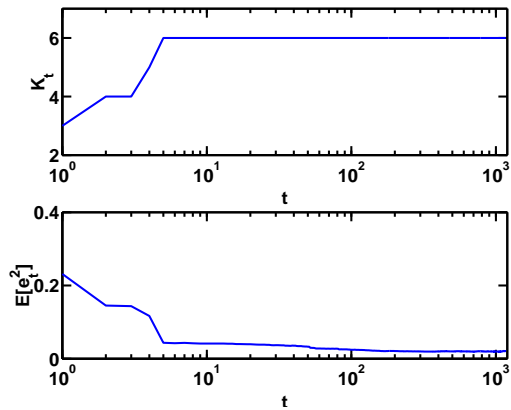}
  \end{center}
  \caption{MOUSSE tracking a static submanifold with $D = 100$ and $d
    = 1$.}
  \label{Fig:static}
\end{figure}

We then study the performance of MOUSSE tracking a static
submanifold.  The dimension of the submanifold is $D = 100$ and the
intrinsic dimension is $d=1$.  Fixing $\theta \in [-2,2]$, we define
$v(\theta) \in \mathbb{R}^D$ according to (\ref{genM}) with $\gamma_t = \gamma = 0.6$ for all $t$. The observation $x_t$ is obtained from
(\ref{data_model}) with noise variance $\sigma^2 =
4\times10^{-4}$. We set $d = 1$ (the assumed intrinsic dimension is identical to the true $d$), $\alpha=0.95$,
$\epsilon=0.1$, $\mu=0.1$, and use PETRELS-FO for subspace tracking. Figure~\ref{Fig:static} demonstrates that MOUSSE is able to track a static submanifold and
reach the steady state quickly from a coarse initialization. 

\subsection{Tracking a slowly time-varying
  submanifold}\label{sec:track}

Next we looking closely at MOUSSE tracking a slowly time-varying submanifold. Consider the submanifold defined in (\ref{slow_manifold}), with $D = 100$ and $d=1$. We set the assumed intrinsic dimension to be identical to the true $d$, choose $\gamma_0=2\times10^{-4}$, $s=1000$, $\mu=0.1$, $\epsilon=0.1$, $\alpha=0.9$ for MOUSSE, and  use PETRELS-FO for subspace tracking. Let
 $40\%$ of the entries missing at random\footnote{The result of the tracking can be found in an illustrative video at
\href{http://nislab.ee.duke.edu/MOUSSE/}{http://nislab.ee.duke.edu/MOUSSE/}}.

\begin{figure}[h!]
  \begin{center}
    \includegraphics[width = 0.4\textwidth]{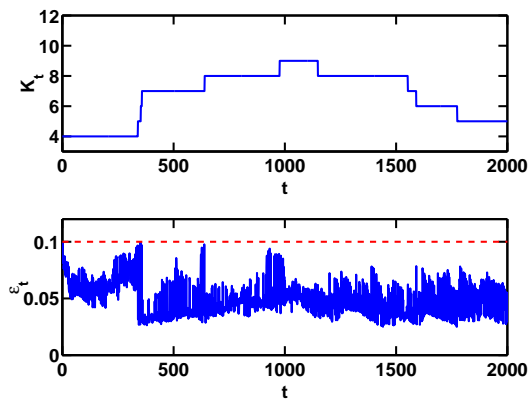}
  \end{center}
  \caption{MOUSSE tracking a slowly time-varying submanifold with $D =
    100$ and $d = 1$. The dashed red line depicts the parameter $\epsilon$ used to control approximation errors in the subset tracking.}
  \label{Fig:changing}
\end{figure}

Snapshots of this video at time $t = 250$ and $t = 1150$ are shown in
Figure~\ref{fig:MOUSSE}.  In this figure, the dashed line corresponds
to the true submanifold, the red lines correspond to the estimated
union of \yao{subsets} by MOUSSE, and the $+$ signs correspond to the past 500
samples, with darker colors corresponding to more recent
observations. From this video, it is clear that we are effectively
tracking the dynamics of the submanifold, and keeping the
representation parsimonious so the number of \yao{subsets} used by our
model is proportional to the curvature of the submanifold. As the
curvature increases and decreases, the number of \yao{subsets} used in our
approximation similarly increases and decreases.  The number of
\yao{subsets} $K_t$ and residuals $e_t$ as a function of time are shown in
Figure~\ref{Fig:changing}. The red line in Figure~\ref{Fig:changing}
corresponds to $\epsilon$. Note that MOUSSE is able to track the
submanifold, in that it can maintain a stable number of leaf nodes in
the approximation and meet the target residual tolerance $\epsilon$.

\subsection{Choice of intrinsic dimension $d$}

In this section, we study the effect of the choice of the intrinsic dimension $d$ in MOUSSE. We generate a chirp-signal, with ambient dimension $D=100$ and signal intrinsic dimension $d_0=2$. Let the two-dimensional parameter be $\theta \triangleq [f_0, \phi]$, with frequency $f_0\in [1, 100]$, and phase $\phi \in [0, 1]$. Define $v(\theta)\in \mathbb{R}^D$ with its $n$-th element
\begin{equation}
[v(\theta)]_n=\sin\left[2\pi (f_0 z_n+\frac{k_t^2}{2}z_n^2+\phi) \right]
\end{equation}
where $z_n=10^{-4}n, n=1,2,\dots,100$, corresponds to regularly-spaced points between 0 and 0.01.  The parameter  $k_t$ controls how fast the submanifold changes and is set according to
\[k_t=\left\{ \begin{array}{ll}   
0.1t, &t=1,2,\dots,1000,\\
200-0.1t, &t=1001,1002,\dots,2000.
 \end{array} \right.\]
Let $40\%$ of the entries be missing at random. 
For MOUSEE, we use PETRELS-FO for tracking. We compare the performance of MOUSSE when $d$ is set within the algorithm to be 1, 2, and 3, so there can be a mismatch between the true intrinsic dimension and the assumed $d$. The parameters of MOUSSE set in these scenarios are: for $d = 1$: $\epsilon=1.5$, $\mu=0.01$, $\alpha=0.95$; for $d=2$: $\epsilon=0.3$, $\mu=0.01$, $\alpha=0.95$;
for $d=3$: $\epsilon=0.3$, $\mu=0.01$, $\alpha=0.95$.

\begin{figure}[h!]
\begin{center}
\subfloat[$d = 1$]{\label{Fig:d1}
\includegraphics[width = .4\linewidth]{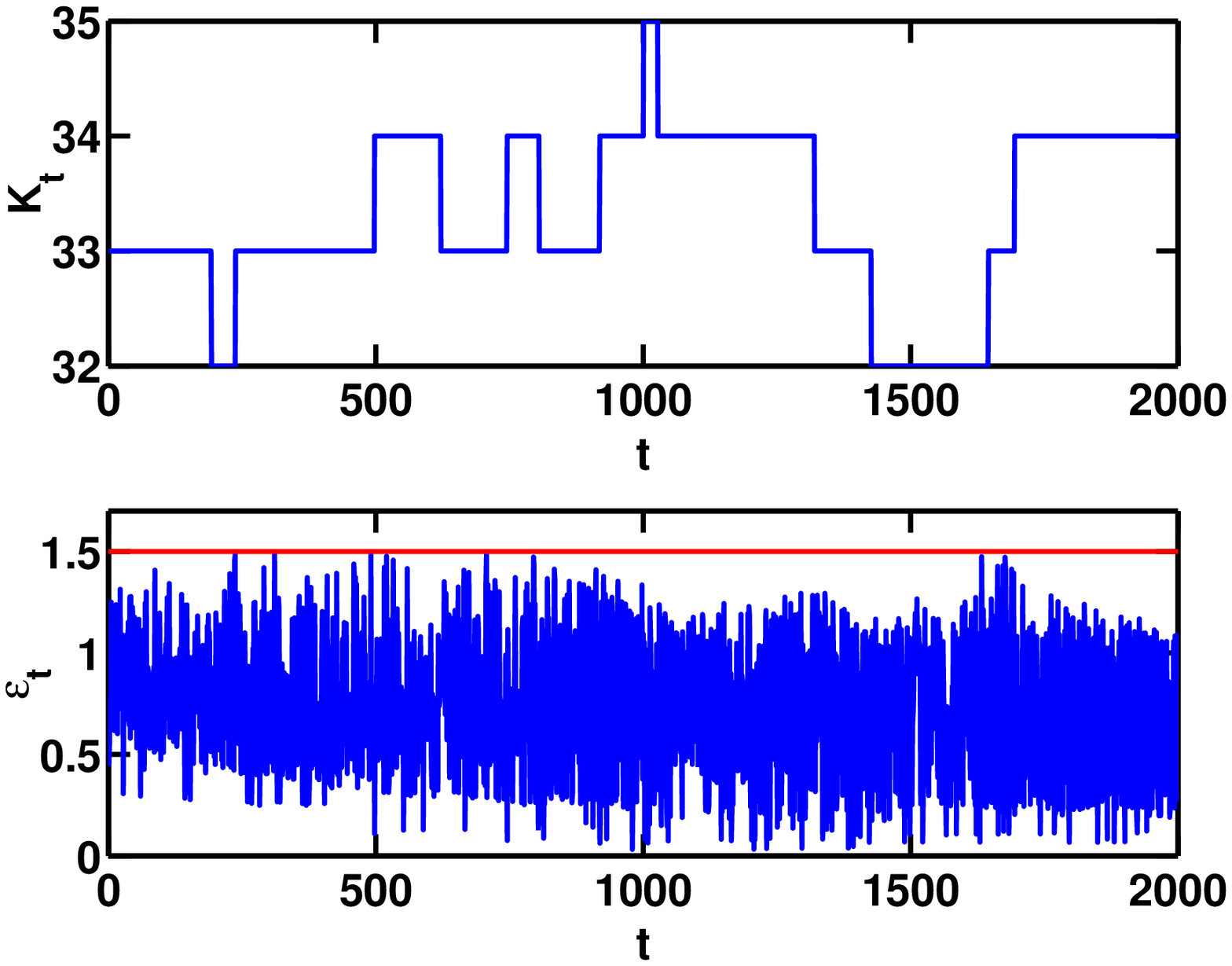}}
\qquad
\subfloat[$d = 2$]
{\label{Fig:d2}
\includegraphics[width = .4\linewidth]{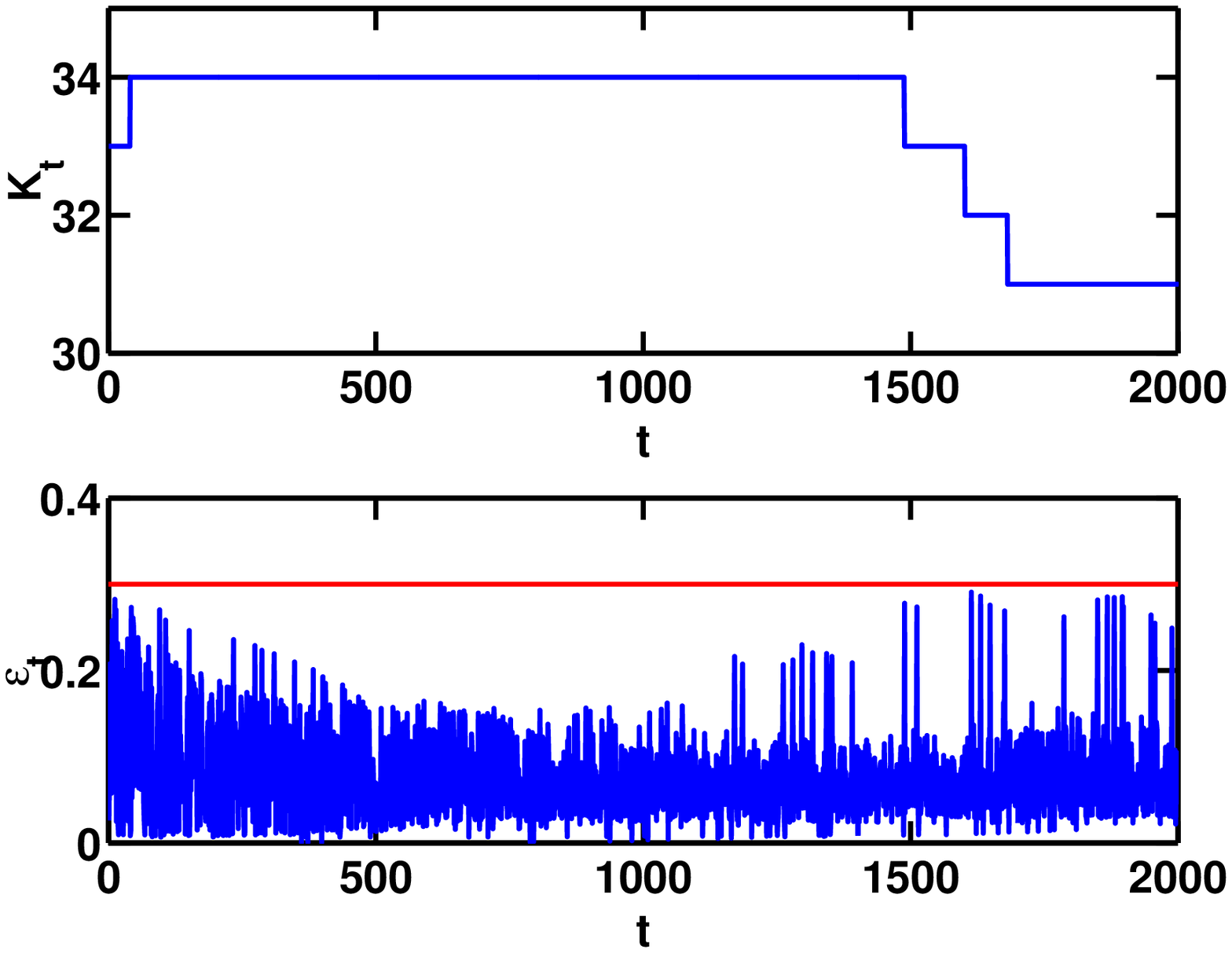}}
\qquad
\subfloat[$d = 3$]{\label{Fig:d3}
\includegraphics[width = .4\linewidth]{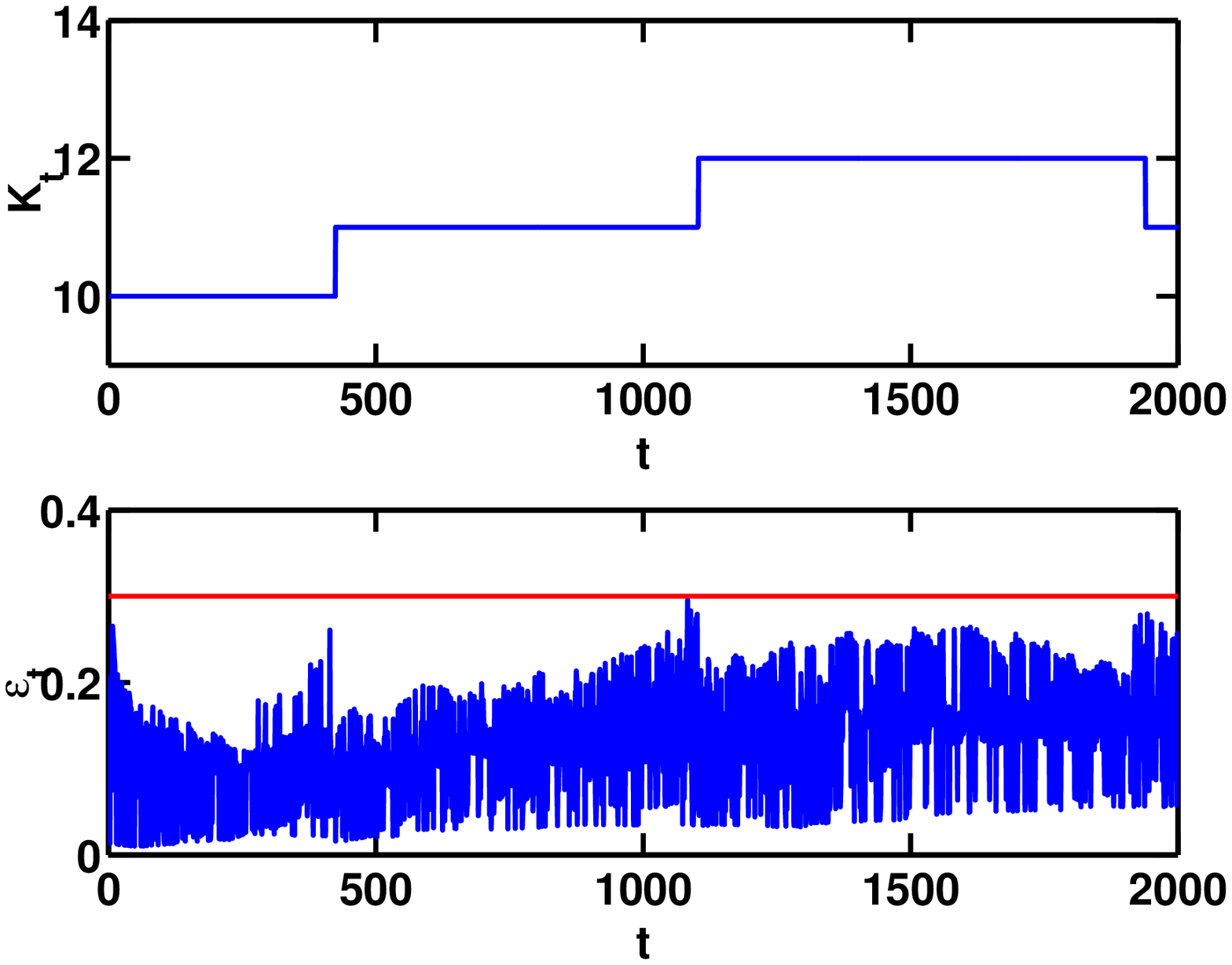}}
\end{center}
\caption{Tracking of MOUSSE using $d = 1$, $d = 2$ and $d = 3$, respectively, when the true intrinsic dimension is 2. Red line corresponds to $\epsilon$.}
\label{Fig:d}
\end{figure}

Fig. \ref{Fig:d} demonstrates that MOUSSE can track the manifold well when the intrinsic dimension is smaller or equal to the assumed $d$. However, if $d$ is chosen to be too small, the errors are significantly larger and we are forced to use a larger error tolerance $\epsilon$. 

\subsection{Change-point detection using MOUSSE}\label{sec:ARL_approx}

\subsubsection{Approximation to ARL} 

\yao{The ARL approximation in (\ref{approx}) assumes $e_t$ is Gaussian distributed. We have shown that $e_t$ is not exactly Gaussian distributed but close to a Gaussian. Hence, we need to numerically verify the accuracy of (\ref{approx}) for $e_t$ generated by MOUSSE.} To simulate ARL of the GLR procedure, we generate $10000$ Monte Carlo (MC) trials, each being a noisy realization of the same slowly time-varying submanifold in (\ref{slow_manifold}). We then apply MOUSSE to track the submanifold, obtain a sequence of residuals $e_t$, apply the GLR change-point detection procedure, and obtain an  ARL numerically. We adopt an exponential approximation in \cite{XieSiegmund2012} to evaluate $\Expect^\infty\{T\}$ efficiently. TABLE~\ref{table:ARL} shows the value of $b$ suggested by theory for different ARLs and the value of $b$'s computed via Monte Carlo are very close. For comparison, we also obtain thresholds for change-point detection when a single subspace tracking using PETRELS-FO is employed.

\begin{table*}[h]
{\small
  \caption{Average run length (ARL) $\Expect^\infty\{T\}$. }
  \begin{center}
    \begin{tabular}{c|c||c|c||c|c||c|c}
      \hline
      \multirow{2}{*}{ARL} & \multirow{2}{*}{$b$} & 
\multicolumn{2}{|c||}{$b$ MC, $0\%$ data
        missing} &  \multicolumn{2}{|c||}{$b$ MC, $20\%$ data
        missing} &  \multicolumn{2}{|c}{$b$ MC, $40\%$ data
        missing} \\ \cline{3-8}
& &  MOUSSE & Single Subspace & MOUSSE & Single Subspace & MOUSSE &
Single Subspace  \\\hline
      1000 & 3.94  &  4.81 &3.90  &4.77 &3.91  & 5.22 & 3.90 
      \\
      5000 & 4.35 & 5.91	&	4.60& 5.66 &4.62  &                                          6.14     &       4.59
      \\
      10000 & 4.52  &  6.38	&	4.91	&		6.02 &		4.91 &                                             6.49 &          4.91 \\ 
      \hline
    \end{tabular}
    \label{table:ARL}
  \end{center}}
\end{table*}

\subsubsection{Comparison of tracking algorithms for MOUSSE}
To estimate the expected detection delay of MOUSSE detecting a change-point, we generate instances where
the parameter $\gamma_t$ in (\ref{genM}) has an abrupt jump $\Delta_\gamma$ at time
$t=200$:  \ben \gamma_t = \left\{\begin{array}{ll}
    0.6-\gamma_0 t & t = 1, 2, \dots, 199, \\
    \gamma_{199}-\Delta_\gamma-\gamma_0 t &t = 200, 201, \dots,400.
  \end{array}\right.
\een 
We apply the GLR procedures based on $e_t$ generated from MOUSSE and single subspace tracking, respectively, and compare the corresponding expected detection delay after $t = 200$. We consider two change-point magnitudes: big ($\Delta_\gamma=0.05$) and small ($\Delta_\gamma=0.03$). The expected detection delays are estimated using 10000 Monte Carlo trials, and are given in
Table ~\ref{table:bigJump}, and Table ~\ref{table:medJump}.
For comparison, we also obtain thresholds for change-point detection when a single subspace tracking using PETRELS-FO is employed. The threshold $b$'s are chosen according to the Monte Carlo
thresholds given in Table~\ref{table:ARL}. For example, for the cell corresponding
to ARL $ =1000$ and $0\%$ missing data in Table~\ref{table:bigJump}
or~\ref{table:medJump}, $b$ should be set as $4.55$ for MOUSSE and
$4.28$ for the single subspace method.
Table~\ref{table:bigJump} and Table ~\ref{table:medJump} demonstrate that change-point detection based on MOUSSE
has a much smaller expected detection delay than that based on single subspace tracking.


\begin{table*}[h]
\small{
  \caption{Detection delay when jump of $\gamma_t$ is
    $\Delta_\gamma = 0.05$. }
  \begin{center}
    \begin{tabular}{c||c|c||c|c||c|c}
      \hline
      \multirow{2}{*}{ARL} &
\multicolumn{2}{|c||}{delay, $0\%$ data missing} &
\multicolumn{2}{|c||}{delay, $20\%$ data missing} &
\multicolumn{2}{|c}{delay, $40\%$ data missing} \\ \cline{2-7}
&  MOUSSE & Single Subspace & MOUSSE & Single Subspace & MOUSSE &
Single Subspace  \\\hline
      1000   &  3.69	&		 91.92		&	 4.02    	 &                90.49                  &                     5.38            &            88.72 
      \\
      5000 &  5.31	 	&	104.02		&	 5.48 	  &              104.23                          &              7.38       &                 105.05
      \\
      10000  & 6.20		&	98.95		&	6.13	   &             101.52                         &                8.21               &         102.99 \\ 
      \hline
    \end{tabular}
    \label{table:bigJump}
  \end{center}}
\end{table*}

\begin{table*}[h]
  \caption{Detection delay when jump of $\gamma_t$ is $\Delta_\gamma =
    0.03$. }
    {\small
  \begin{center}
    \begin{tabular}{c||c|c||c|c||c|c}
      \hline
      \multirow{2}{*}{ARL} &
\multicolumn{2}{|c||}{delay, $0\%$ data missing} &
\multicolumn{2}{|c||}{delay, $20\%$ data missing} &
\multicolumn{2}{|c}{delay, $40\%$ data missing} \\ \cline{2-7}
&  MOUSSE & Single Subspace & MOUSSE & Single Subspace & MOUSSE &
Single Subspace  \\\hline
      1000   & 2.30		&	 54.17		&	 2.39   	   &              52.82                       &                   2.78             &            51.53 
      \\
      5000 &    2.71	 	&	 80.61		&	 2.76 	   &              78.29                         &                 3.35           &              75.47 
      \\
      10000  & 2.91		&	 90.87	&		 2.94	&                 88.30               &                           3.62       &                 86.48\\ 
      \hline
    \end{tabular}
    \label{table:medJump}
  \end{center}}
\end{table*}

\subsection{Real data}


\subsubsection{Solar flare detection}

\begin{figure*}[h!!!]
  \begin{center}
    \subfloat[Snapshot of original SDO data at
    $t=227$]{\label{Fig:original}
      \includegraphics[width = 0.34\textwidth]{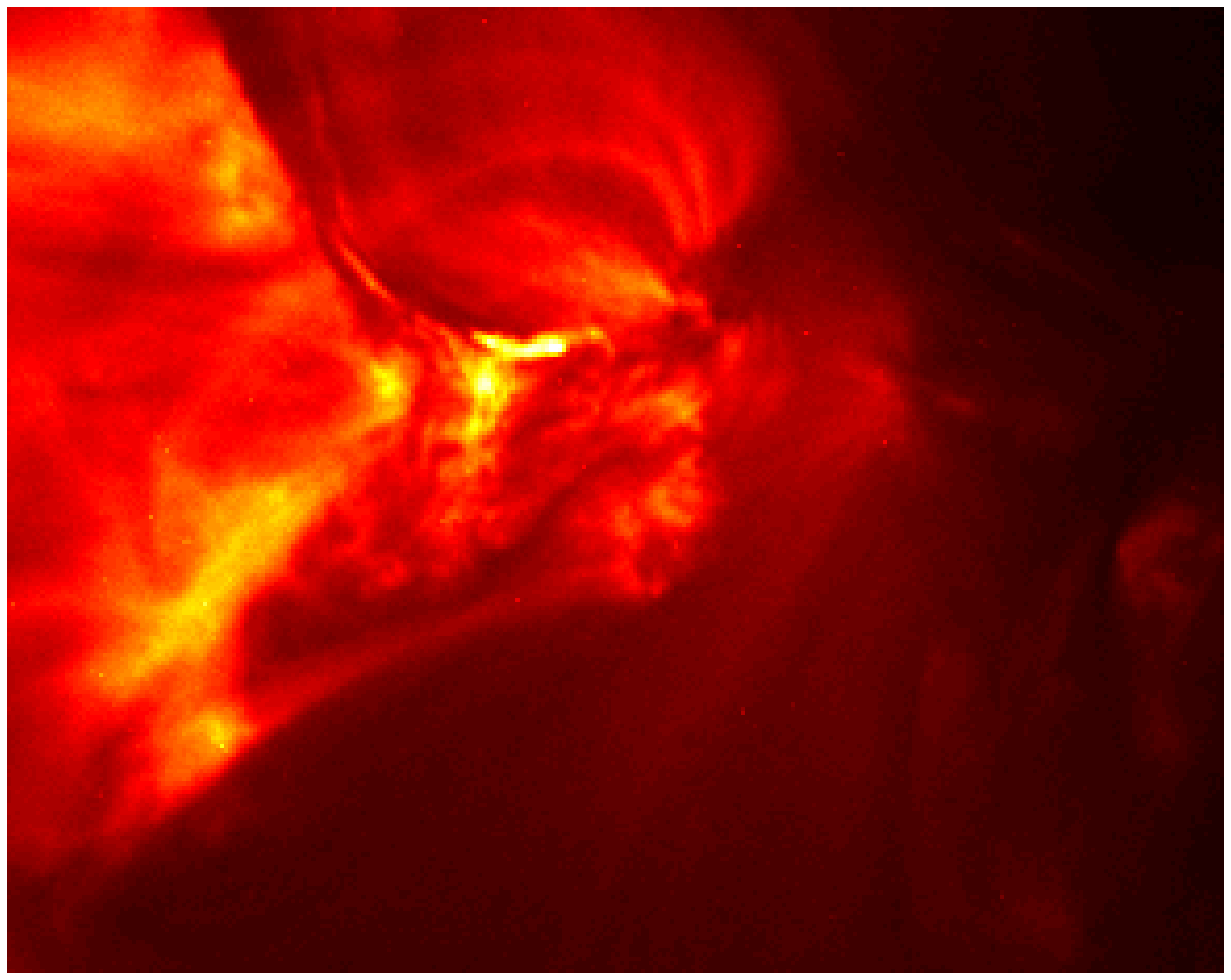}}
    \subfloat[MOUSSE residual map at $t=227$]{\label{Fig:residual}
      \includegraphics[width = 0.34\textwidth]{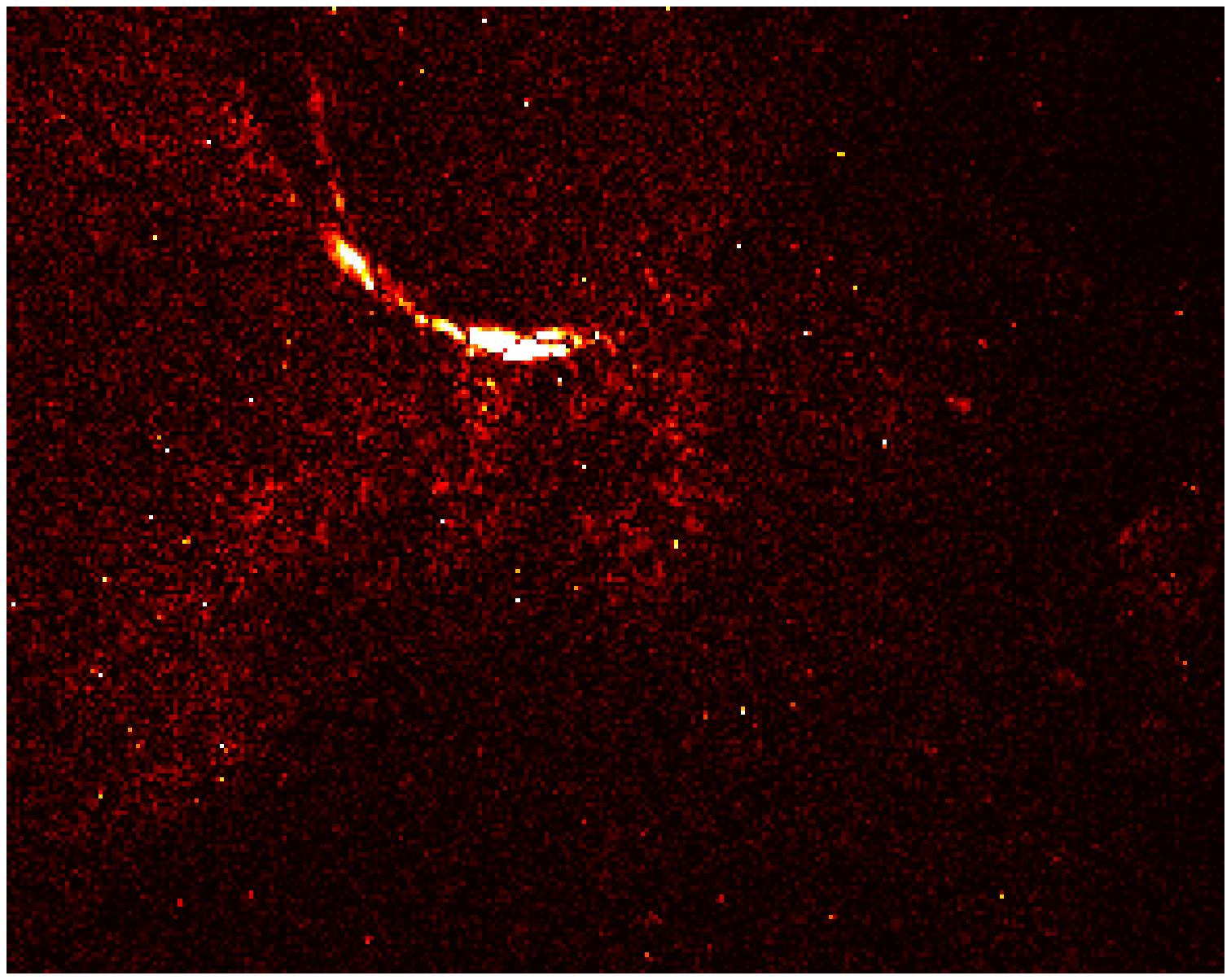}}
    \subfloat[Single subspace tracking residual map at
    $t=227$]{\label{Fig:residual1sub}
      \includegraphics[width = 0.34\textwidth]{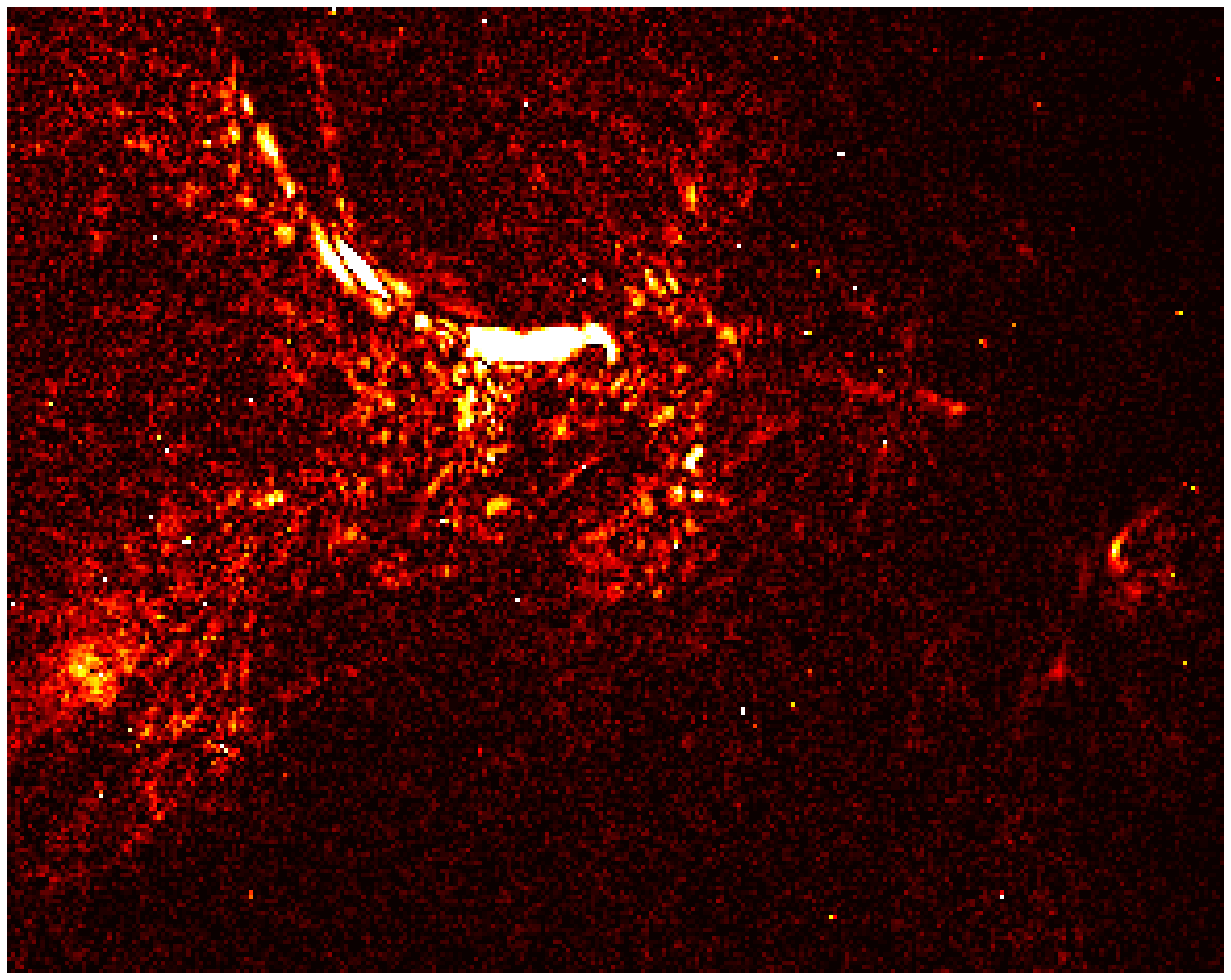}}\\
    \subfloat[$e_t$ from MOUSSE]{\label{Fig:solar_et}
      \includegraphics[width = .24\textwidth]{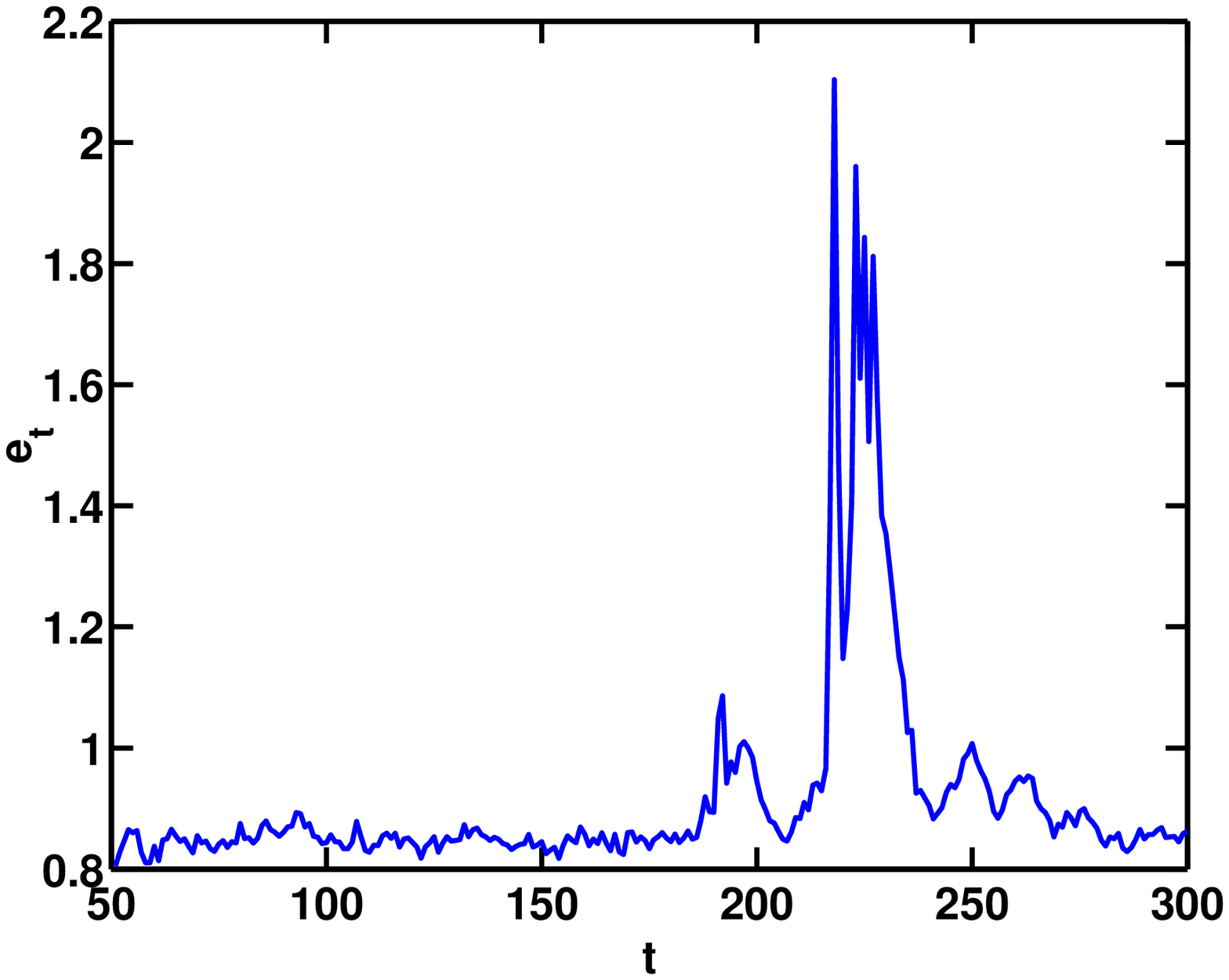}}
    \subfloat[GLR stats from MOUSSE]{\label{Fig:solar_stats}
      \includegraphics[width = .24\textwidth]{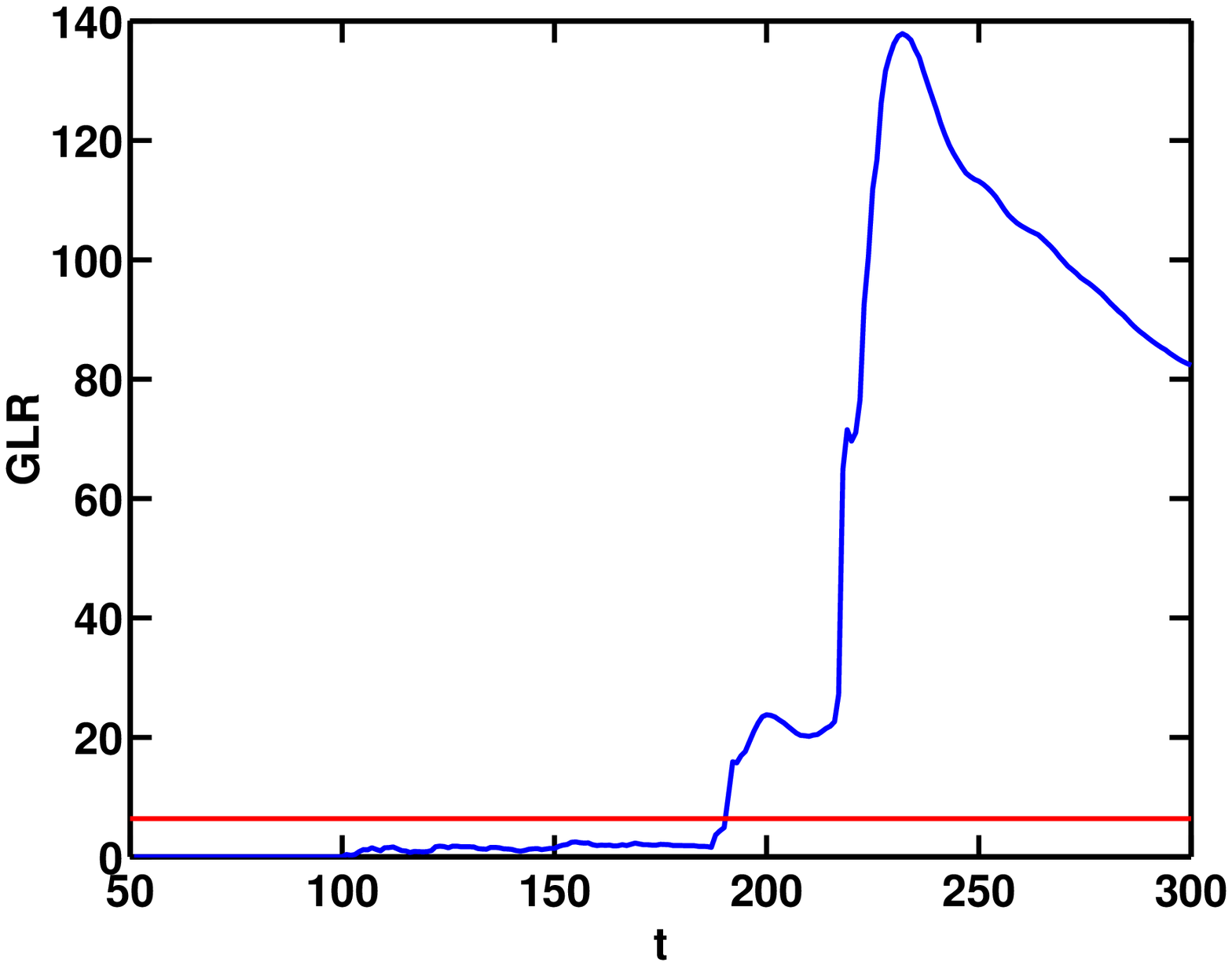}}
    \subfloat[$e_t$ from single subspace
    tracking]{\label{Fig:solar_et_1sub}
      \includegraphics[width = .24\textwidth]{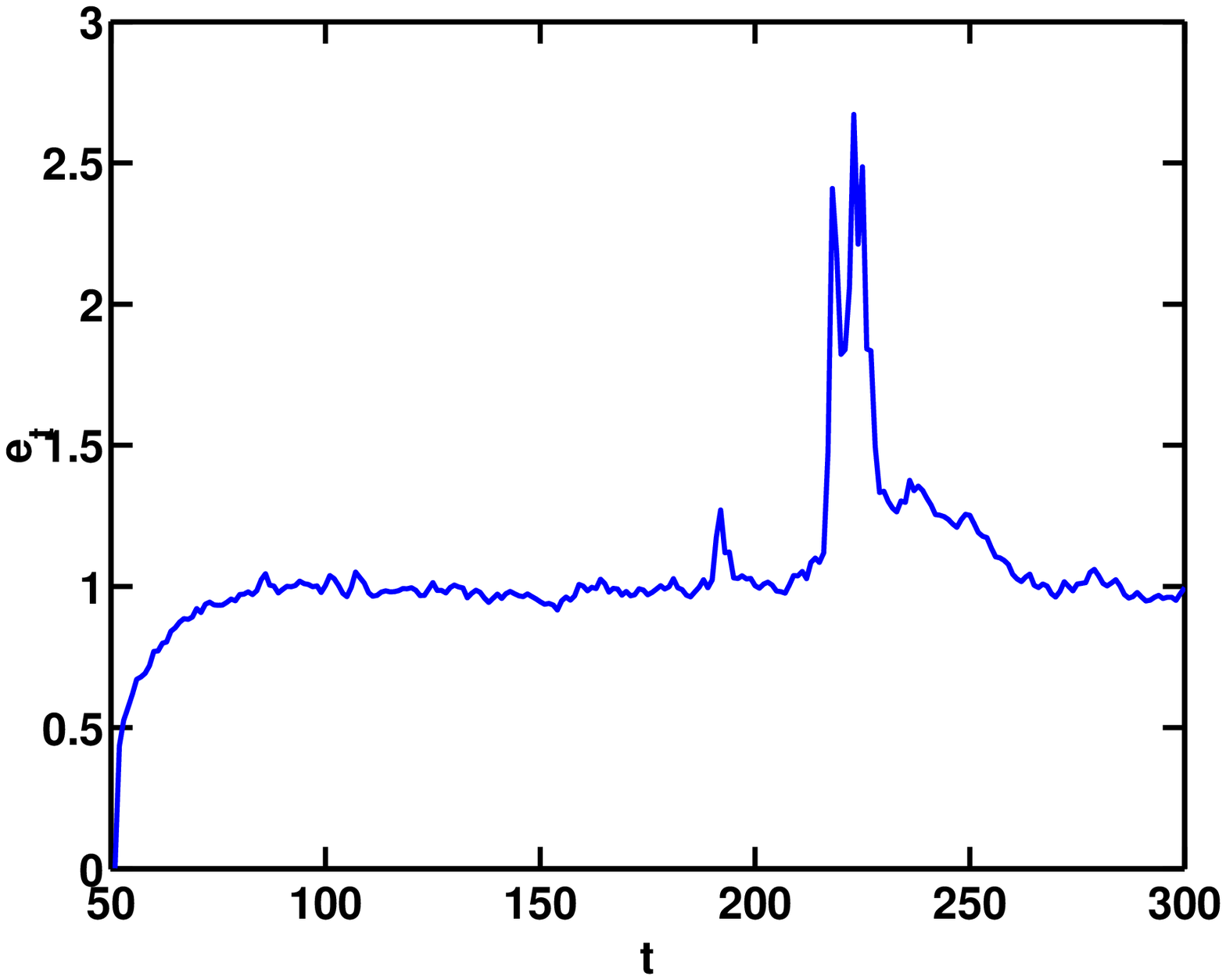}}
    \subfloat[GLR stats from single subspace tracking
    ]{\label{Fig:solar_stats_1sub}
      \includegraphics[width = .24\textwidth]{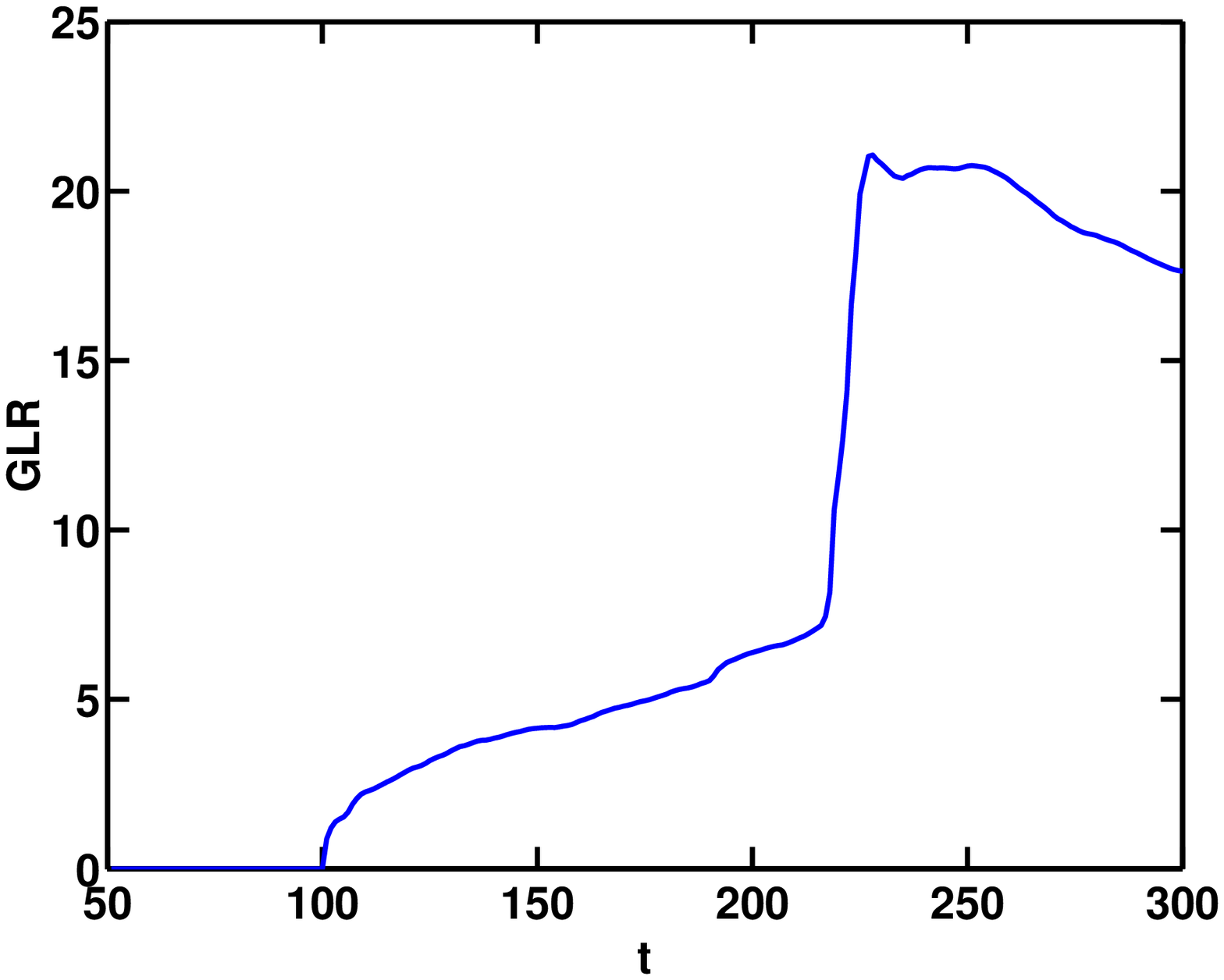}}
  \end{center}
  \caption{Detection of solar flare at $t = 227$: (a) snapshot of
    original SDO data at $t = 227$; (b) MOUSSE residual $\hat{e}_t$,
    which clearly identifies an outburst of solar flare; (c) single
    subspace tracking residual $\hat{e}_t$, which gives a poor
    indication of the flare; (d) $e_t$ for MOUSSE which peaks near
    the flare around $t = 227$; (e) the GLR statistic for MOUSSE;
    (f) $e_t$ for single subspace tracking; (g) the GLR statistic
    for single subspace tracking. Using a single subspace gives much
    less reliable estimates of significant changes in the statistics
    of the frames. }
  \label{Fig:solar}
\end{figure*}

\yao{We first consider a video from the Solar Data Observatory, which demonstrates an abrupt emergence of a solar flare} \footnote{The video can be found at \href{http://nislab.ee.duke.edu/MOUSSE/}{http://nislab.ee.duke.edu/MOUSSE/}. The Solar Object Locator for the original data is  SOL2011-04-30T21-45-49L061C108}. \yao{We also display a residual map defined as:
\begin{equation}
  \hat{e}_t \triangleq (I-U_{j^*, k^*, t}U_{j^*, k^*, t}^\#)(x_t- c_{j^*, k^*, t}),\label{err_vector}
\end{equation}
which is useful to localize the solar flare. Here $(j^*, k^*)$ denotes the index of the minimum distance subset.} The frame is of size $232\times292$ pixels, which result in $D = 67744$ dimensional streaming data. In this video, the normal states are slowly drifting solar flares, and the anomaly is a much brighter transient solar flare. A frame from this dataset during a solar flare around $t = 200$ is shown in Figure~\ref{Fig:original}.  In the original images,
the background solar images have bright spots with slowly
changing shape, which makes detection based on simple background
subtraction incapable of detecting small transient flares.


To ease parameter tuning, we scale the pixel intensities by a factor of $10^{-4}$, so the range of data is consistent with our simulated data experiments. 
The parameters for this example are \yao{$d = 1$}, $\epsilon=0.3$, $\mu=0.3$, and $\alpha=0.85$. Figure~\ref{Fig:solar} demonstrates that MOUSSE can not
only detect the emergence of a solar flares, but also localize the
flare by presenting $\hat{e}_t$, and these tasks are accomplished far
more effectively with MOUSSE (even with $d=1$) than with a single
subspace. Note that with  single subspace tracking, $e_t$  is not a stationary timeseries prior to the flare and
thus poorly suited for change-point detection. In
contrast, with our approach, with $K_t$ around $10$, the underlying
manifold structure is better tracked and thus yields more stable $e_t$ before the change-point and significant change in $e_t$  when the change-point occurs.

\subsubsection{Identity theft detection}

Our second real data example is related to automatic identity theft
detection. The basic idea is that consumers have typical spending
patterns which change abruptly after identity theft. Banks would like
to identify these changes as quickly as possible without triggering
numerous false alarms. To test MOUSSE on this high-dimensional
changepoint detection problem, we examined the E-commerce transaction
history of people in a dataset used for a 2008 UCSD data mining competition\footnote{
Data available at
\url{http://www.cs.purdue.edu/commugrate/data_access/all_data_sets_more.php?search_fd0=20}.}.
For each person in this dataset, there is a timeseries of
transactions. For each transaction we have a 31-dimensional
real-valued feature vector and a label of whether the transaction is
``good'' (0) or ``bad'' (1). The full dataset was generated for a
generic anomaly detection problem, so it generally is not appropriate
for our setting. However, some of these transaction timeseries show a
clear changepoint in the labels, and we applied MOUSSE to these
timeseries. In particular, we use MOUSSE to track the 31-dimensional
feature vector and detect a changepoint, and compare this with the
``ground truth'' changepoint in the label timeseries.  In calculating
the GLR statistic, we estimate the $\mu_0$ and $\sigma_0$ of
equation \ref{det_proc} from $e_1,\dots,e_{20}$.  After $t=20$, every
time the GLR statistic exceeds the threshold $b$ and an changepoint
is detected, we ``reset'' the GLR to only consider $e_t$ {\em after}
the most recently detected changepoint. This allows us to detect
multiple change-points in a timeseries.

\begin{figure}[h!]
  \begin{center}
    \subfloat[Obtained from MOUSSE]{\label{Fig:MOUSSE credit}
    \includegraphics[width = 0.25\textwidth]{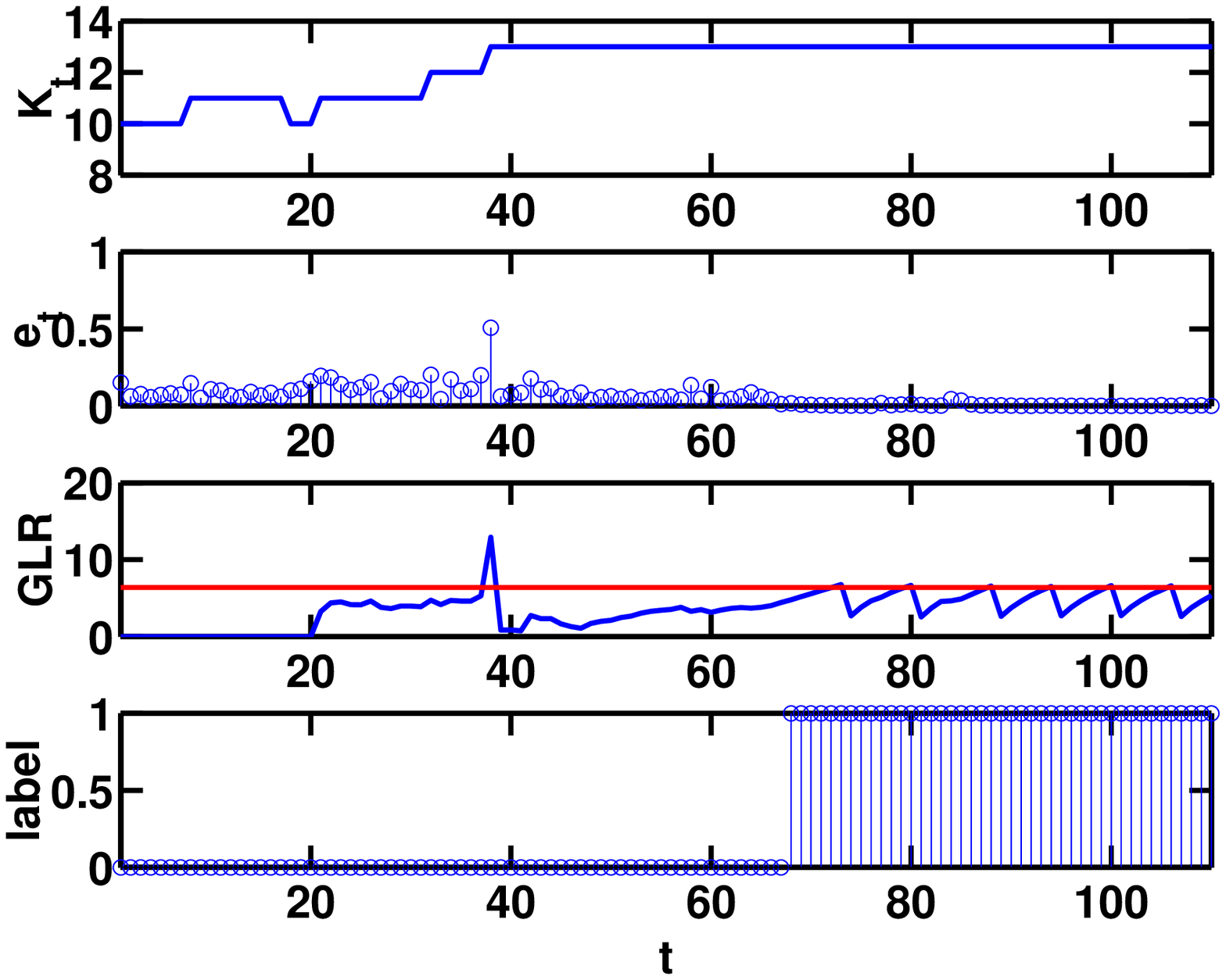}}
    \subfloat[Visualization of time-varying Attributes]{\label{Fig:data visualization}
    \includegraphics[width = 0.23\textwidth]{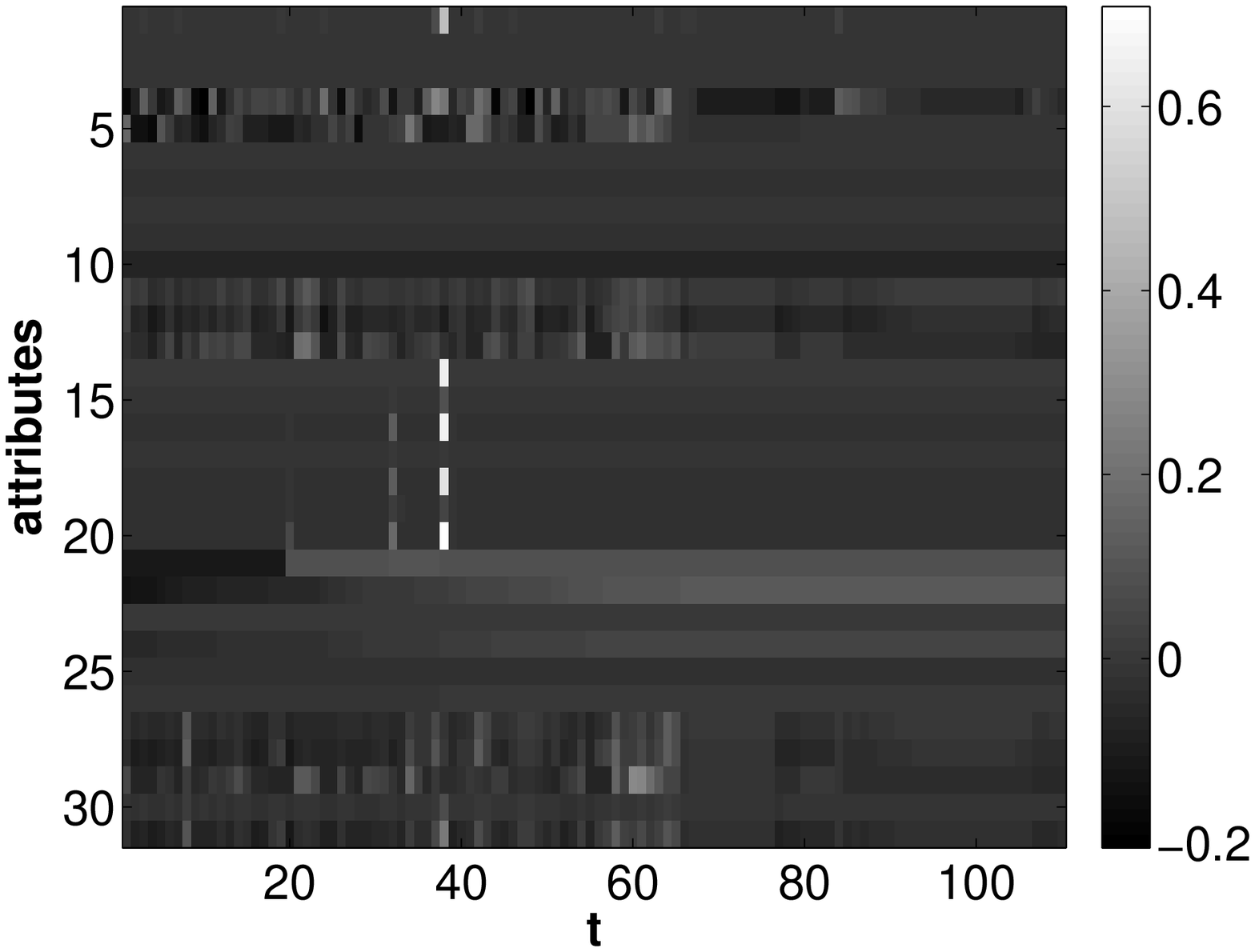}}
  \end{center}
  \caption{Credit card user data experiments. (a) From top to bottom:
    number of leaf nodes used by MOUSSE; $e_t$; GLR statistic (solid
    blue line) and theoretical threshold $b$ corresponding to ARL =
    10000 (dashed red line); ground truth label. Note that the GLR
    statistic has a false alarm due to an outlier at $t = 38$, and it
    starts increasing at $t = 70$ and frequently hits the threshold
    afterwards due to the changepoint at $t=70$. In this case GLR
    catches both the outlier and the changepoint. (b) Demonstration of
    the time-varying $x_t$ (user attributes): each column corresponds
    to the 31-dimensional attribute vector at a given time. The white
    spots correspond to the outlier at time $t=38$.}
  \label{Fig:credit}
\end{figure}

The effect of our procedure for one person's transaction history is
displayed in Figure~\ref{Fig:credit}. We first see that MOUSSE
accurately detects a temporally isolated outlier transaction at
$t=38$, after which the GLR is reset. After this, while MOUSSE does
not generate particularly large spikes in $e_t$, the associated GLR
statistic shows a marked increase near $t=70$ and hits the threshold
at $t=72$ \yao{(the threshold corresponds to the Monte Carlo threshold for ARL = 10000 in Table \ref{table:ARL})} when the labels (not used by MOUSSE) change from 0 (good)
to 1 (bad). After this the GLR is repeatedly reset and repeatedly
detects the change in the statistics of $e_t$ from the initial
stationary process.

\section{Conclusions}
This paper describes a novel multiscale method for online tracking of
high-dimensional data on a low-dimensional submanifold, and using the
tracking residuals to perform fast and robust change-point
detection. Change-point detection is an important subset of anomaly
detection problems due to the ever-increasing volume of streaming data
which must be efficiently prioritized and analyzed. The multiscale
structure at the heart of our method is based on a geometric
multiresolution analysis which facilitates low-complexity
piecewise-linear approximations to a manifold. The multiscale
structure allows for fast updates of the manifold estimate and
flexible approximations which can adapt to the changing curvature of a
dynamic submanifold. These ideas have the potential to play an
important role in analyzing large volumes of streaming data which
arise in remote sensing, credit monitoring, and network traffic
analysis.

While the algorithm proposed in this paper has been focused on unions
of \yao{subsets}, an important open question is whether similar techniques
could be efficiently adopted based on sparse covariance matrix
selection \cite{ABE08,KakadeShamirSridharan2010}. The resulting
approximation space may no longer correspond to a low-dimensional
submanifold, but such structures provide good representations of
high-dimensional data in many settings, and our future work includes
tracking the evolution of a mixture of such structures. Issues related
to non-Gaussian observation models, inverse problem settings,
dynamical models, and optimal selection of the statistic used for
change-point detection (\ie alternatives to $e_t$, as considered in
\cite{XieSiegmund2012JSM}) all pose additional interesting open
problems.


\appendices

\section{Optimality of estimate for $c$}\label{sub:c}
We assume
that there is complete data, and we restrict our approximation to a
single subspace so that $K_t = 1$. Assume the mean and covariance
matrix of the data are given by $c^\star$ and $\Sigma^\star$,
respectively. Assume the covariance matrix has low-rank structure:
$\Sigma^\star = \mbox{diag}\{\lambda_1^\star, \ldots,
\lambda_D^\star\}$ with $\lambda_m= \delta^\star$ for $m = d+1,
\ldots, D$.

When there is only one subspace and the data are complete, the cost
function (\ref{M_opt}) without the penalty term becomes 
\begin{equation}
 \min_{U,
  c} \sum_{i=1}^t \alpha^{t-i} \|(I-UU\transpose)(x_i-c)\|^2.  \label{M_opt_1}
  \end{equation} 
Recall that the online update for $c_t$ is given by $c_{t+1} = \alpha c_{t}
+ (1-\alpha) x_t$, with initialization $c_0$. We can prove that this online estimate for $c$ is optimal in the following sense:
\begin{thm}\label{thm_opt_c}
Assume $c^\star_t$ minimizes (\ref{M_opt_1}) at time $t$, $0\leq \alpha < 1$, and the initialization is bounded $\|c_0\|^2 < \infty$. Then as $t\rightarrow \infty$, $\|c_t - c^\star_t\|^2 \rightarrow 0$ in probability. Moreover, assume $x_t$'s are i.i.d. with $\mathbb{E}\{x_t\} = c^\star$, then $\mathbb{E}\{c_t\} \rightarrow c^\star$, i.e., the estimate is asymptotically unbiased.
\end{thm}

{\em Proof:} Recall that the online estimate for $c_t$ is given by $c_{t+1} = \alpha c_{t}
+ (1-\alpha) x_t$. Hence,
\[
c_t = (1-\alpha)\sum_{i=1}^{t}\alpha^{t-i} x_i + \alpha^{t} c_0,
\]
where the term $\alpha^{t} c_0$ is a bias introduced by initial
condition $c_0$. 
Let 
\begin{equation}
  \bar{x}_t = \frac{1-\alpha}{1-\alpha^t}\sum_{i=1}^t \alpha^{t-i}x_i, \quad 
  S = \sum_{i=1}^t \alpha^{t-i}(x_i - \bar{x}_t)(x_i - \bar{x}_t)\transpose.
\end{equation}
By
expanding $\|(I-UU\transpose)(x_i-c)\|^2 =
\|(I-UU\transpose)(x_i - \bar{x}_t + \bar{x}_t - c)\|^2$, and using the fact that $(I-UU\transpose)^2 = I-UU\transpose$, we can write the
cost function of (\ref{M_opt_1}) as
\begin{equation}
\begin{split}
&\sum_{i=1}^t \alpha^{t-i} \|(I-UU\transpose)(x_i-c)\|^2 \\
=&   \sum_{i=1}^t \alpha^{t-i} (x_i - \bar{x}_t)\transpose  (I-UU\transpose) (x_i - \bar{x}_t)\\
&+  \sum_{i=1}^t \alpha^{t-i} (\bar{x}_t - c)\transpose  (I-UU\transpose) (\bar{x}_t - c) \\
&+ 2 \sum_{i=1}^t \alpha^{t-i}  (\bar{x}_t -c)\transpose (I-UU\transpose) (x_i - \bar{x}_t) .
\end{split}
\label{intermediate}
\end{equation}
Since $\bar{x}_t$ and $c$ are both independent of $i$, the last term in (\ref{intermediate}) can be re-written and is equal to zero by the choices of $\bar{x}_t$ and $S$:
\begin{equation}
\begin{split}
&\sum_{i=1}^t \alpha^{t-i}  (\bar{x}_t-c)\transpose (I-UU\transpose) (x_i - \bar{x}_t) \\
= &  (\bar{x}_t -c)\transpose (I-UU\transpose) \sum_{i=1}^t \alpha^{t-i} (x_i - \bar{x}_t)\\
=& (\bar{x}_t - c)\transpose (I-UU\transpose) \left(\sum_{i=1}^t \alpha^{t-i} x_i - \bar{x}_t \sum_{i=1}^t \alpha^{t-i}\right) = 0,
\end{split}
\label{zero}
\end{equation}
since $\sum_{i=1}^t \alpha^{t-i} = (1-\alpha^t) /(1-\alpha)$. Using the fact that $\tr(AB) = \tr(BA)$ for two matrix $A$ and $B$, together with (\ref{zero}), the cost function (\ref{intermediate}) becomes
\begin{equation}
  \tr[(I-UU\transpose)S]
  + 
  \frac{1-\alpha^t}{1-\alpha}(\bar{x} - c)\transpose(I - UU\transpose) (\bar{x} - c),
  \label{new_cost}
\end{equation}
where the first term does not depend on $c$.  Since the second term in (\ref{new_cost}) is quadratic in $c$, it is
minimized by choosing $c = \bar{x}_t$. Denote this optimal $c$ at time $t$ by $c^\star_t$.

Hence 
\begin{equation}
\|c_t^\star -c_t \|^2 \leq
\left\|\frac{\alpha^t}{1-\alpha^t} (1-\alpha)\sum_{i=1}^t \alpha^{t-i} x_i\right\|^2
+ \alpha^t \left\|c_0
\right\|^2. \label{diff}
\end{equation} 
Recall that  $c^\star$ denote the true mean: $\mathbb{E}\{x_t\}  = c^\star$.
As $t\rightarrow \infty$, $\sum_{i=1}^t \alpha^{t-i} x_i \rightarrow \frac{1}{1-\alpha} c^\star$ in probability, the first term in the upper bound (\ref{diff}) tends to 0 in probability. Given bounded $\|c_0\|^2$, the second term in (\ref{diff}) also tends to 0. Hence our online-estimate $c_t$ is asymptotically optimal in that it minimizes (\ref{M_opt_1}). 
Also, $c_t$ is asymptotically unbiased, since $\Expect\{c_t\}
\rightarrow (1-\alpha) \cdot\frac{1}{1-\alpha} c^\star =
c^\star$.

\section{Consistency of estimates of $\Lambda^\star$ and
  $\delta^\star$} \label{sec:consistency}

We assume
that there is complete data, and we restrict our approximation to a
single subspace so that $K_t = 1$.
In the following, we show that  if we have
correct $U = U^\star$, then for each sample $x_t$, its projection
$[\beta_t]_m$ is an unbiased estimator for $\lambda_m^\star$, and
$\|x_{t, \perp}\|^2$ is an unbiased estimator for $\sum_{m=d+1}^D
\lambda_m^\star$. First note \ben
\begin{split}
  \Expect\{|[\beta_t]_m|^2 \} &= \Expect\{ [\textsf{e}_m\transpose
  U\transpose(x_t - c)]^2\}
  = \textsf{e}_m\transpose U \transpose \Sigma^\star U \textsf{e}_m \\
  &= \lambda_m^\star [U]_m\transpose [U]_m = \lambda_m^\star,
\end{split}
\een for $m = 1, \ldots, d$, where $\textsf{e}_m$ denotes the $m$-th
row of an identity matrix.
We also have that \ben
\begin{split}
  \Expect\{\|x_{t, \perp}\|^2\}  &= \Expect\{\|(I-UU\transpose)(x_t - c)\|^2\}\\
  &= \tr\{(I-UU\transpose)\Sigma^\star(I-UU\transpose)
  \}\\
  &= \sum_{m=d+1}^D \lambda_m^\star.
\end{split}
\label{perp_expect}
\een Then from the MOUSSE update equations, as $t\rightarrow \infty$
\ben
\begin{split}
  \Expect\{\lambda^{(m)}_t\} &= \Expect\{(1-\alpha)\sum_{i=1}^t
  \alpha^{t-i} |[\beta_t]_m|^2 + \alpha^t \lambda_0^{(m)}\} \rightarrow \lambda_m^\star,
\end{split}
\label{lambda_est}
\een for $m = 1, \ldots, d$ and \ben
\begin{split}
  \Expect\{\delta_t\} &= \Expect\{(1-\alpha)\sum_{i=1}^t \alpha^{t-i} \|x_{t, \perp}\|^2/(D-d) + \alpha^t \delta_0\} \\
  &\rightarrow \frac{1}{D-d}\sum_{m=d+1}^D \lambda_m^\star =
  \delta^\star.
\end{split}
\label{delta_est}
\een Hence our estimators for $\lambda_m^\star$ and $\delta^\star$ are
asymptotically unbiased. 
%

\section{Proof of Theorem~\ref{thm_missing}}
\label{app1}

\begin{proof}
  From (\ref{data_model}) and (\ref{beta}) we have \ben \beta =
  U\transpose(v-c) + U\transpose w, \label{beta_expr} \een Note that
  $U\transpose w$ is zero-mean Gaussian random vector with covariance
  matrix $\sigma^2 U\transpose U = \sigma^2 I$.

  Next we consider the missing data case. Recall
  $\cP_{\Omega}\in\mathbb{R}^{|\Omega|\times D}$ is a projection
  matrix. Define $w_\Omega = \cP_{\Omega}w$. From (\ref{beta}) we have
  \ben \beta_{\Omega} & = U_\Omega^\# (v_\Omega-c_\Omega) +
  U_\Omega^\# w_\Omega \een
  Suppose in (\ref{data_model}) we write $v - c = p + q$, with $p \in
  \mathcal{S}$ and $q \in \mathcal{S}^\perp$, where $\Set^\perp$
  denotes the orthogonal subspace of $\mathcal{S}$. Hence, $p =
  UU\transpose(v - c)$ and $q = (I - UU\transpose)(v - c)$. Let
  $p_\Omega = \cP_{\Omega}p$, $q_\Omega = \cP_{\Omega}q$. Hence,
  $v_\Omega - c_\Omega = p_\Omega + q_\Omega$.
  Note that
  \begin{align}
    U_\Omega^\# p_\Omega
    =&  (U_\Omega\transpose U_\Omega)^{-1} U_\Omega\transpose \cP_\Omega UU\transpose (v-c) \\
    =& (U_\Omega\transpose U_\Omega)^{-1} U_\Omega\transpose U_\Omega U\transpose (v-c) \\
    =& U\transpose(v-c).
  \end{align}
  So \[\beta_\Omega = U\transpose (v-c) + U_\Omega^\# q_\Omega +
  U_\Omega^\# w_\Omega.\] Hence \ben
  \begin{split}
    \|\beta_{\Omega} - \beta\|^2 
    &\leq 2 \|  U_\Omega^\# q_\Omega \|^2 +2 \| U_\Omega^\# w_\Omega - U\transpose w\|^2 \\
    &= 2\|(U_\Omega\transpose U_\Omega)^{-1}U_\Omega\transpose
    q_\Omega\|^2 \\
    &+ 2\| [(U_\Omega\transpose U_\Omega)^{-1}
    U_\Omega\transpose \cP_\Omega - U\transpose] w \|^2
  \end{split}\nonumber
  \een
  We will bound these two terms separately.

  First, note that \ben \|(U_\Omega\transpose
  U_\Omega)^{-1}U_\Omega\transpose q_\Omega\|^2 &
  \leq\|(U_\Omega\transpose U_\Omega)^{-1}\|_2^2 \|U_\Omega\transpose
  q_\Omega\|^2
  \label{bound_Uq}
  \een where $\|A\|_2$ denotes the spectral norm of matrix $A$. Using
  [Lemma 2] in \cite{BalzanoRechtNowak2011}, we have that with
  probability $1-\varepsilon$, if $|\Omega| \geq
  \frac{8}{3}d\textsf{coh}(U)\log(2d/\varepsilon)$,
  \[
  \|U_\Omega\transpose q_\Omega\|^2 \leq (1+\theta)^2
  \frac{|\Omega|}{D}\frac{d}{D}\textsf{coh}(U)\|q\|^2,
  \]
  where $\theta = \sqrt{2\frac{\max_{n=1}^D
      |[q]_n|^2}{\|q\|^2}\log(1/\varepsilon)}$. Using [Lemma 3] in
  \cite{BalzanoRechtNowak2011} we have that provided that $0< \ell <
  1$, with probability at least $1-\varepsilon$, \ben
  \|(U_\Omega\transpose U_\Omega)^{-1}\|_2 \leq
  \frac{D}{(1-\ell)|\Omega|}.
  \label{L2_norm}
  \een Combine these with (\ref{bound_Uq}), we have that with
  probability $1-2\varepsilon$, \ben \|(U_\Omega\transpose
  U_\Omega)^{-1}U_\Omega\transpose q_\Omega\|^2 \leq
  \frac{(1+\theta)^2 }{(1-\ell)^2 } \cdot\frac{d}{|\Omega|}\cdot
  \textsf{coh}(U)\|q\|^2.
  \label{residual_bound}
  \een

  Next we examine the noise term.
  Define \[\tilde{w} =[(U_\Omega\transpose U_\Omega)^{-1}
  U_\Omega\transpose \cP_\Omega - U\transpose] w,\] which is a
  zero-mean Gaussian random vector with covariance
  matrix \[\Gamma=\sigma^2(U_\Omega\transpose U_\Omega)^{-1} -
  \sigma^2I,\] where we have used the fact that $\cP_\Omega
  \cP_\Omega\transpose = I$. Hence we bound the tail of the noise
  power using Markov inequality:
  \begin{equation}
    \mathbb{P}(\|\tilde{w}\|^2 > 2\tau^2\sigma^2)
    \leq e^{-\tau} \mathbb{E}\{e^{\|\tilde{w}\|^2/(2\tau\sigma^2)}\}\leq 2 D e^{-\tau}
    \label{inequal}
  \end{equation}
  provided that $\tau$ is sufficiently large such that the maximum
  eigenvalue is smaller than $\tau$:
  $\lambda_{\max}((U_\Omega\transpose U_\Omega)^{-1})< \tau$, \ie
  $\tau > D/[(1-\ell)|\Omega|],$ by noting that
  $\lambda_{\max}((U_\Omega\transpose U_\Omega)^{-1}) =
  \|(U_\Omega\transpose U_\Omega)^{-1}\|_2$. The last equality in
  (\ref{inequal}) is because, under such condition: \ben
  \begin{split}
   \mathbb{E}\{e^{\|\tilde{w}\|^2/(2\tau\sigma^2)}\} 
    &= \int e^{\|x\|^2/(2\tau\sigma^2)} (2\pi)^{-D/2} |\Gamma|^{-1/2}
    e^{-\frac{1}{2}x\transpose\Gamma^{-1}x} dx \\
    & = (2\pi)^{-D/2} |\Gamma|^{-1/2} \int e^{-\frac{1}{2}x\transpose(\Gamma^{-1}-\tau^{-1}\sigma^{-2} I)x} dx\\
    & =  |\Gamma|^{-1/2} |\Gamma^{-1}-\tau^{-1}\sigma^{-2}I|^{-1/2}  \\
    &=  |I - \tau^{-1}\sigma^{-2}\Gamma|^{-1/2}\\
    &= |(1+1/\tau)I - \tau^{-1}(U_\Omega\transpose U_\Omega)^{-1}|^{-1/2}\\
    &\leq D[(1+1/\tau)- \tau^{-1}\|(U_\Omega \transpose U_\Omega)^{-1}\|_2]^{-1/2}\\
    &\leq D\left[(1+1/\tau) - \frac{D}{\tau(1-\ell)|\Omega|}
    \right]^{-1/2}\\
    & = D\left[1 - \frac{1}{\tau}(\frac{D}{(1-\ell)|\Omega|}-1)
    \right]^{-1/2}
  \end{split}\nonumber
  \een In the last inequality, we have used (\ref{L2_norm}). Note that
  $D/[(1-\ell)|\Omega|] > 1$, and the upper bound in
  (\ref{noise_bound_large}) is smaller than $2D$ if $\tau >
  \frac{4}{3}(\frac{D}{(1-\ell)|\Omega|}-1)$ or $\tau >
  \frac{4}{3}\frac{D}{(1-\ell)|\Omega|}$. Now we set $2De^{-\tau} =
  \varepsilon$, if $\varepsilon$ is sufficiently small such that
  $\log(2D/\varepsilon) >
  \frac{4}{3}\frac{D}{(1-\ell)|\Omega|}$. Hence we have when $|\Omega|
  > \frac{4}{3}\frac{D}{(1-\ell)\log(2D/\varepsilon)}$,
  $\|\tilde{w}\|^2 <
  \frac{32}{9}\frac{D^2\sigma^2}{(1-\ell)^2|\Omega|^2}$ with
  probability $1-\varepsilon$.  Finally, combining
  (\ref{residual_bound}) and the noise bound above,
  we obtain the
  statement in Theorem~\ref{thm_missing}.

%
%
%

\end{proof}

\bibliographystyle{IEEEbib} \bibliography{MOUSSE}

\end{document}